\pdfoutput=1

\documentclass[11pt]{article}

\usepackage[]{acl}
\usepackage{natbib}

\usepackage{times}
\usepackage{latexsym}

\usepackage[T1]{fontenc}

\usepackage[utf8]{inputenc}
\usepackage{longtable}
\usepackage{microtype}

\usepackage{inconsolata}

\usepackage{graphicx}

\usepackage[utf8]{inputenc} 
\usepackage{url}            
\usepackage{booktabs}       
\usepackage{amsfonts}       
\usepackage{xcolor}         
\usepackage{pifont}
\usepackage{graphicx}
\usepackage{amsmath}
\usepackage{array}
\usepackage{xspace}
\usepackage{makecell}
\usepackage{subcaption}
\usepackage{float}
\usepackage{multirow}
\usepackage{enumitem}

\usepackage{cleveref}
\crefname{table}{Table}{}
\crefname{figure}{Figure}{}
\crefname{section}{\S}{}
\crefname{appendix}{Appendix}{}

\newcommand{\cmark}{\ding{51}}%
\newcommand{\xmark}{\ding{55}}

\newcommand{\ngrammatch}[0]{\textsc{ngram-match}\xspace}

\newcommand{\openai}[0]{\textsc{ngram-match}\xspace}
\newcommand{\ngram}[0]{\textsc{token-match}\xspace}
\newcommand{\union}[0]{\textsc{token-extend}\xspace}
\newcommand{\skipgram}[0]{\textsc{longest-match}\xspace}

\newcommand{\plotmethodname}[0]{\texttt{ConTAM}\xspace}

\title{Evaluation data contamination in LLMs: how do we measure it and (when) does it matter?\vspace{5mm}}

\author{Aaditya K. Singh\thanks{Shared first authorship} \\
  University College London\\
  \texttt{aaditya.singh.21@ucl.ac.uk} \\\And
  Muhammed Yusuf Kocyigit\textsuperscript{*}\thanks{Work done while at Meta} \\
  Boston University\\
  \texttt{kocyigit@bu.edu} \\
  \And
  Andrew Poulton\textsuperscript{$\dagger$}\\
  Cohere\\
  \texttt{andrewpoulton@cohere.com}\\
  \AND
  David Esiobu\\
  Meta\\\And
  Maria Lomeli\\
  Meta\\\And
  Gergely Szilvasy\\
  Meta\\\AND
  Dieuwke Hupkes \\
  Meta \\
  \texttt{dieuwkehupkes@meta.com}
  }

\begin{document}
\maketitle


\begin{abstract}
Evaluation data contamination, the inadvertent mixing of samples from evaluation benchmarks into pre-training corpora, constitutes a recently growing and important concern in the field of evaluating large language models (LLMs).
The resulting `training on the test set' makes it difficult to interpret evaluation benchmark scores, and an active area of research studies its effects.
However, while evaluation data contamination is easily understood intuitively, it is surprisingly difficult to define precisely which examples should be considered contaminated and, consequently, to what extent contamination inflates the corresponding benchmark scores.
In this paper, we propose that these questions should be addressed together and that contamination metrics can be assessed based on whether the examples they mark contaminated indeed give models an undue advantage.
We propose a novel analysis method called \plotmethodname, and show -- in a large scale survey of existing and novel $n$-gram based contamination metrics across 13 benchmarks and 7 models from 2 different families -- that \plotmethodname can be used to better understand evaluation data contamination as well as its effects on benchmark scores.
We find that contamination may have a much larger effect than reported in recent LLM releases and that there are differences in the extent to which models at different scale are impacted by contamination.
Furthermore, we find that considering only the \emph{longest contaminated substring} generally provides a better signal than considering a union of all contaminated substrings, as was common practice in previous studies, and that doing model and benchmark specific threshold analysis greatly increases the specificity of the results.
Lastly, we investigate the impact of various hyperparameter choices of contamination studies, finding that -- among other things -- both using larger values of $n$ and disregarding contaminated strings that are infrequent in the pre-training data lead to many false negatives.
With \plotmethodname, we provide a method to empirically ground evaluation data contamination metrics in downstream effects as well as measure their magnitude.
With our exploration, we shed light on how evaluation data contamination can impact LLMs and provide insight into the various considerations important when doing contamination analysis.
We end our paper by discussing these in more detail and providing concrete suggestions for future work.
\end{abstract}

\section{Introduction}
\label{sec:introduction}

As large language models (LLMs) improve, their evaluation becomes increasingly challenging.
A frequent concern in this context \citep[i.a.]{conda-2024-data,sainz-etal-2023-nlp,jacovi-etal-2023-stop,elazar2024whats} is that samples from evaluation benchmarks sometimes (partially) appear in the data used to train that same model, also known as \emph{evaluation data contamination}.
Such inadvertent `training on the test set' makes it difficult to interpret scores on evaluation benchmarks: does a better score reflect an improvement on the ability that the benchmark intends to measure or does it indicate that a model succesfully memorised the benchmark data?

Answering this seemingly simple question is far from straightforward.
Even setting aside practical issues such as the sheer size of pre-training corpora and the fact that they are not often publicly available, it is surprisingly difficult to define when precisely a sample in an evaluation benchmark should be considered contaminated.
Does the entire sample need to be verbatim present in the training data, or should partial appearance also count?
If so, how do we define partial appearance, and how do we `threshold' it?
In addition to these measurement questions, there is the intertwined question of how much contamination really impacts benchmark scores.
If an example occurs verbatim once in a 1.4T token training corpus, does this in fact give the model an unfair advantage?

In this paper, we propose a novel methodology for assessing contamination metrics that we call the \textit{Contamination Threshold Analysis Method} (\plotmethodname).
\plotmethodname is centered around the concept of \emph{Estimated Performance Gain (EPG)}, and addresses the former two questions jointly.
Motivated by the idea that an adequate metric of contamination returns the set of samples for which a model in fact has an undue benefit, it compares metrics based on the measurable impact on benchmark performance of the samples the metrics flag as contaminated.
In doing so, it simultaneously provides handholds to make empirical choices regarding contamination metrics and provides information on answers the question of the extent to which contamination inflates benchmark scores.

We use \plotmethodname to compare \emph{four main contamination metrics} -- three from the literature and one novel -- that define contamination in terms of overlapping $n$-gram spans (termed ``matches'') between a sample in the evaluation benchmark and the pre-training corpus.
We compare these metrics across \emph{5 parameter settings}, 
on \emph{13 benchmarks}, for \emph{7 models} of various sizes trained from \emph{2 pre-training corpora}.
Our comparisons illustrate how \plotmethodname can help provide insights into how evaluation data contamination impacts model performance, help distinguish contamination metrics, and select hyper-parameters for those metrics.
We conduct extensive experimentation across these metrics and corresponding hyperparameter choices.
Our main conclusions are:
\begin{itemize}[itemsep=0.1em]
    \item The \textbf{impact of evaluation data contamination has been underestimated in many prominent LLM releases}, likely because of false negatives in the chosen contamination metrics (\cref{tab:comp_method_z}, Figures \cref{fig:skip_epg_pct_contam_all,fig:skip_pct_contam_epg_best,fig:skip_epg_pct_contam_dist})
    \item While there is no true \emph{one-size-fits-all} approach to contamination detection, \textbf{using the \emph{longest} contaminated substring rather than a union of all matches works better} across the board, adequately detecting contamination in cases where no other metric did (Figures \cref{fig:main_comp_method,fig:appx_comp_method})
    \item For virtually all benchmarks we considered, \textbf{smaller $n$ is better} and \textbf{even one occurrence in the pre-training data matters}: both using values of $n$ larger than 8 and setting a minimal count higher than one for occurrence in the pre-training corpus leads to false negatives (Figure \cref{fig:piqa_false_neg,fig:impact_mincount});
    \item \textbf{The impact of contamination changes with scale}, in cases where there still is performance to be gained, larger models are better able to leverage contamination than smaller models (Figure \cref{fig:scale,fig:scaling_llama_all,fig:scaling_pythia_all});
    \item Consequently, to find the most adequate contamination metric, it is important to do \textbf{model-specific threshold selection} (\cref{fig:false_positives}).
\end{itemize}

\noindent With our work, we aim to contribute to an informed discussion about evaluation data contamination and its effects, as well as provide methodology for both researchers and practitioners to explore this question further and contrast the effect of different contamination metrics and hyper-parameters as newer pre-training corpora \citep{soldaini2024dolmaopencorpustrillion, penedo2024finewebdatasetsdecantingweb} and benchmarks \citep{laurent2024labbenchmeasuringcapabilitieslanguage} are released. 

\paragraph{Outline} In the remainder of this paper, we first review earlier work on contamination detection, and consider their results and limitations (\cref{sec:related}).
Next, we describe the various contamination metrics we investigate in more detail and list the benchmarks, models and pre-training corpora we use for our study (\cref{sec:methods}), followed by a description of our main methodology for analysis (\cref{sec:analysis_methods}).
After that, in our main results section (\cref{sec:results}), we first report the results of an overall comparison between all contamination metrics (\cref{subsec:comparing_methods}).
Both numerically and qualitatively, we show how different contamination metrics provide different signal across datasets, and we show that taking the longest matching $n$-gram, rather than a union of all $n$-grams, detects the most meaningful EPG across datasets and models.
Next, in \cref{subsec:found_contamination}, we discuss how much contamination is detected across various benchmarks (\cref{subsec:how_contaminated}), how much that impacts benchmark scores for various models (\cref{subsec:how_impactful}) and how that depends on model scale (\cref{subsec:scaling}).
Consecutively, in \cref{sec:analysis}, we investigate how various parameters that frequently occur in contamination metrics impact the results.
Specifically, we consider the impact of $n$ (\cref{sec:analysis_n}), the impact of the mismatch budget (\cref{subsec:skip_budget}) and the frequency of matches in the pre-training corpus (\cref{sec:analysis_frequency}). 
We wrap up with a conclusion in which summarise our results and provide concrete recommendations for practitioners (\cref{sec:conclusion}) and a discussion in which we consider the limitations of our work (\cref{sec:discussion}).

\section{Related work}
\label{sec:related}

One of the primary reasons evaluation data is considered problematic is that it can render evaluations unreliable and -- therefore -- comparisons unfair.
Recently, the literature around evaluation data contamination has grown, with some works mentioning how it is a more prevalent problem \citep{sainz2024datacontaminationreport2024, balloccu2024leakcheatrepeatdata} in current evaluations, especially as LLM outputs become popularly shared on the internet \citep{natolambert2024contam}. 
While most practitioners attempt to \textit{decontaminate} pre-training corpora before training, such approaches are never perfect \citep{brown2020fewshot,yang2023rethinkingbenchmarkcontaminationlanguage} thus emphasising a need for better understanding of the impact of evaluation data contamination.
Here, we discuss the three main ways of tackling this question.

\subsection{Causal contamination analysis}

Perhaps the most direct way to measure the impact of contamination in training data, causal contamination analysis involves specifically training models on contaminated data and measuring the increase in benchmark performance. 
\citet{jiang2024investigatingdatacontaminationpretraining} perform an extensive study using this method, finding that contamination can lead to modest improvements. 
However, their study is limited to considering small sized models (124M parameters), as pre-training from scratch is expensive. 
Other works, instead, consider finetuning pre-trained models and demonstrate significant effects of contamination when training on benchmark samples such as websites containing HellaSwag extracts \citep{geminiteam2024geminifamilyhighlycapable}, rephrased benchmark samples \citep{yang2023rethinkingbenchmarkcontaminationlanguage}, or even reformatted samples \citep[e.g.\ JSON versions of HumanEval questions,][]{geminiteam2024gemini15unlockingmultimodal}. 
The finetuning approach scales more easily to larger models but does not provide as a clear a signal on the effects of contamination interspersed in a pre-training corpus. 
Furthermore, such causal methods continue to be expensive to compute, as even in the finetuning case they would require finetuning a large model on various datasets with various perturbations.

\subsection{Post-hoc contamination analysis}

An alternative to the causal analysis, post-hoc contamination analyses focus on identifying possible inflation in benchmark scores after a model has been trained. 
Such approaches focus on identifying samples from evaluation benchmarks that were contaminated in the pre-training data, subsetting the benchmark to a ``clean'' subset, and measuring the difference in performance between the complete benchmark and this clean subset. 
Unlike causal methods, these approaches can only give a correlational notion of contamination, but benefit from not needing to do any training of a model (and thus are easier to use on public artifacts). 
Furthermore, they are more amenable to studying the effect of various notions of contamination within a single experiment.

Prior work has used various definitions to define what makes an example contaminated.
\citet{brown2020fewshot} consider any sample with at least one $n$-gram match (where $n$ ranges from 8 to 13 tokens, depending on the benchmark) in pre-training data as contaminated. 
\citet{openai2023gpt4}, instead, check three random 50-character strings, indicating a lack of consistency even by the same group of researchers. 
Other large model practitioners use yet other approaches, with \citet{chowdhery2022palm} counting samples with over 70\% of their $n$-grams occurring in pre-training data as contaminated,  while \citet{touvron2023llama2} consider a skip-gram approach in which small mismatches are allowed.
Generally, choices for contamination metrics vary, and to our knowledge, no systematic study has been done to see the effects of various `hyperparameters' in these metrics. 
In our work, we consider a wide range of contamination metrics and compare them through the lens of performance impact to give benchmark and pre-training corpus specific recommendations. 

\subsection{Memorisation in large language models}\label{subsec:related_memorisation}

Memorisation in large language models \citep{carlini2023quantifying,hartmann2023sokmemorizationgeneralpurposelarge} is closely related to the notion of contamination, as it similarly involves regurgitation of data seen in pre-training. 
Specifically, the literature at the intersection of memorisation and contamination tends to focus on approximating contamination without access to pre-training data, for instance via membership inference attacks \citep[e.g.][]{mireshghallah2022quantifyingprivacyrisksmasked}. 
\citet{golchin2023time} check if the model can complete examples from the dataset with relevant context given to the model; 
\citet{shi2023detecting} use the average probability of the k-least frequent tokens from a sentence to measure if the model has seen that sentence before; and \citet{Golchin2023DataCQ} create slightly different versions of evaluation set examples using GPT-4 and check if the model can find the original example with high precision. 
Finally, \citet{oren2023proving} measure if a model has seen large pieces of the dataset together by checking if the model gives higher probability to the canonical order of the original dataset compared to a randomly shuffled order of examples.
While these methods are valuable, current methods that do not actually search the pre-training data have been shown to be ineffective for accurately detecting contamination \citep{duan2024membership}. 
In our work, we opted to focus on post-hoc contamination analyses that look for matches in pre-training data that we were able to access for the purposes of this study.

\section{Methods}
\label{sec:methods}

For our study, we consider four different contamination metrics across various different hyperparameters.
In this section, we discuss those metrics (\cref{subsec:metrics}), and we list the models and benchmarks we investigate (\cref{sec:methods:models} and \cref{sec:methods:benchmarks}, respectively).
After, in \cref{sec:analysis_methods}, we introduce and discuss our methods for comparing and analysing these metrics.

\begin{table*}[t]
    \centering
    \begin{tabular}{lccccc}
        \toprule
        \textit{Method}    & \textit{Normalisation} & \emph{Search method} & \emph{Scoring} & \emph{Join matches} \\ 
    \midrule
         \openai &   \cmark & $n$-gram & $n$-gram & \cmark \\ 
         \ngram &   \cmark & $n$-gram & token & \cmark \\ 
         \union &   \xmark & $n$-gram \& extend & token & \cmark \\ 
         \skipgram &   \xmark  & $n$-gram \& extend & token & \xmark \\ 
         \bottomrule
    \end{tabular}
    \caption{\textbf{Overview of the differentiating features for contamination metrics.} The four contamination metrics studied in this paper differ on four key features. \emph{Normalisation} refers to whether the method requires lowercasing and punctuation removal before computing matches, the \emph{search method} specifies whether a mismatch budget is allowed, \emph{scoring} signifies whether scoring is based on number of tokens contaminated or number of $n$-grams contaminatiod, and \emph{join matches} indicates whether all matches are joined, or only a single match is counted.}\label{table:contamination_methods}
\end{table*}

\subsection{Quantifying contamination}
\label{subsec:metrics}

We consider four contamination metrics.
Three are motivated by prior work \citep{brown2020fewshot, touvron2023llama2,chowdhery2022palm}, and one comprises a novel metric that we empirically found useful.\footnote{
A detailed overview of which methods are used in which previous work, along with their main results, is provided in \cref{tab:lit_overview_table}.
}
Each of these metrics specifies how contaminated an evaluation sample is through a \textit{contamination score} between 0 and 1, where 0 indicates the sample is not contaminated at all, and 1 that the sample is maximally contaminated.
The contamination scores can then be used to divide evaluation benchmarks in `clean' and `contaminated' partitions, depending on whether the score is surpassing a predetermined \emph{contamination threshold}, prescribing what contamination score a sample should exceed to be considered as contaminated.
Below, we elaborate how the contamination score is defined for the methods under consideration, a brief overview of their main differences is provided in \cref{table:contamination_methods}.

\paragraph{\ngram}
A common method used to assign contaminated scores, first popularised by \citet{brown2020fewshot}, is to consider how many tokens in an evaluation sample are part of an $n$-gram occurring in the pre-training corpus.
\citet{brown2020fewshot} normalise both the pre-training corpus and evaluation sample by converting to lowercase and removing punctuation before looking for matches.
They then define the contamination score as the fraction of tokens in the evaluation sample that occurs in a contaminated $n$-gram.
\citet{brown2020fewshot} used a single $n$ for each dataset (with a value between 8 and 13) and considered all $n$-grams that occurred at least once in the pre-training corpus.
In our study, we vary both the value of $n$ and the minimum frequency of the $n$-gram (referred to with the term $mincount$) as hyperparameters of the method.
We consider values of $n\in\{8, 10, 13, 20\}$ and $mincount\in\{1, 5, 10, 20, 100\}$. 
To be consistent with other methods, we consider $n$-grams in terms of tokens rather than words (i.e.\ an 8-gram is 8 tokens long, rather than 8 words long).

\paragraph{\ngrammatch}
The \ngrammatch method, used by \citet{chowdhery2022palm}, is very similar to \ngram, except that the score is calculated as a fraction of the contaminated \emph{$n$-grams} in the evaluation sample, rather than contaminated tokens. \footnote{For example, a contaminated 9-token span would be counted as two contaminated 8-token spans (and thus have twice the contamination score) of a contaminated 8-token span (for the same sample), assuming $n=8$. This differs from \ngram, which would scale with the number of contaminated tokens (9 vs 8).}
Otherwise, the hyperparameters for this method are the same as the hyperparameters for \ngram, as is the normalisation.
\citet{chowdhery2022palm} use a value of $n$=8 and $mincount$=1; in our study, we consider the same values as we do for \ngram.

\paragraph{\union}
First used by \citet{touvron2023llama2}, the \union metric notably differs from the prior two methods in that it allows for a $skip\_budget$, which allows for substitution mismatches between a contaminated token string in the evaluation sample and the pre-training corpus. 
Using a match-then-extend algorithm, \union identifies all contaminated spans of length at least $n=8$ tokens, and extends them to their longest possible match with a possible $skip\_budget$, then calculates the contamination score as a fraction of tokens in the evaluation sample that were marked contaminated (similar to \ngram). 
Because the skip budget can be used to account for differces in formatting as well as other differences, \union does not use normalisation.
\citet{touvron2023llama2} consider values of $n\in\{10, 20, 30, 40, 50\}$, and a skip budget of $4$.
Here, we consider $n\in\{8, 10, 13, 20\}$, and $skip\_budget\in\{0,1,2,3,4,5\}$.

\paragraph{\skipgram}
Building off \union, we introduce a fourth metric, that we call \skipgram.
The key difference between \union and \skipgram, is that the latter, instead of searching for all contaminated token spans, only considers the \textit{longest} token span (again, with a possible $skip\_budget$). 
The score is then calculated as the fraction of tokens that is part of this longest match. 
Intuitively, this method avoids assigning high contamination scores to samples for which many tokens are contaminated when the contamination consists of different $n$-grams stemming from different places in the pre-training corpus.
Furthermore, it mitigates the issue where templated strings -- e.g.\ ``is to the left of the red ball'' \citep{benchekroun2023worldsense} or ``to make every square inch of the'' \citep{gsm8k} -- confound the contamination scores.

\subsection{Models}
\label{sec:methods:models}

In our experiments, we consider two families of decoder-only autoregressive large language models: Llama 1 \citep{touvron2023llama} and Pythia \citep{biderman2023pythiasuiteanalyzinglarge}, trained on two different pre-training corpora.
Within each family, we consider multiple models to get a notion of generalisability across scale.
For the Llama series, we consider all four model sizes (7B, 13B, 33B, and 65B); for Pythia, we consider the three largest sizes (1.4B, 6.9B, and 12B). 

\subsection{Benchmarks}
\label{sec:methods:benchmarks}

In our analyses, we consider 13 frequently used benchmarks covering a broad range of capabilities.
Specifically, we include COPA \citep{gordon2011copa}, HellaSwag \citep{zellers2019hellaswag}, SIQA \citep{siqa}, PIQA \citep{bisk2019piqa}, Big Bench Hard \citep{bbh}), TriviaQA \citep{joshi-etal-2017-triviaqa}, GSM8k \citep{gsm8k}, MATH \citep{math}, HumanEval \citep{humaneval}, MBPP \citep{mbpp}, DM Contest \citep{dm_contest}), MMLU \citep{mmlu}, and Natural Questions \citep{naturalquestions}. 
We perform contamination analyses for each method and model on each of these benchmarks. 
Some of these benchmarks (MMLU, HellaSwag, PIQA, SIQA, and COPA) are choice benchmarks -- in those cases, we concatenate the correct choice to the prompt when calculating contamination. 
More elaborate descriptions of each benchmark are provided in \cref{appx:example_prompts}.

\section{Analysis methods}
\label{sec:analysis_methods}

Without a pre-determined ground truth on what constitutes contamination, comparing contamination metrics quantitatively is challenging.
Here, we propose to compare metrics by grounding them in the effect that the contamination they detect has on benchmark scores, building on the intuition that a useful contamination metric is one that detects contamination that has an measurable effect on evaluation outcomes.
To quantify this effect, we consider the \emph{estimated performance gain} (EPG) that the detected contamination results in, further discussed in \cref{subsec:epg}.
To compare metrics across hyperparameters and thresholds, we use two analysis methods, one numerical method relying on \emph{z-scores} (\cref{subsec:zscore}) and one qualitative method that considers the behaviour of each metric with changing thresholds (\cref{subsec:qualitative_analysis}).

\subsection{Estimated performance gain (EPG)}
\label{subsec:epg}
The first key concept in our analysis is the EPG that results from the contamination that a metric detects.\footnote{Under a different name, this metric was also used by previous studies, starting from \citet{brown2020fewshot}.}
The EPG for a contamination metric and corresponding contamination threshold is model-dependent and is defined as the difference of the model's performance on the entire benchmark and the subsample of that benchmark marked by the method as uncontaminated or `clean'.
As such, it provides an empirical measure of the effects of contamination while accounting for the size of the contaminated set -- if the clean and contaminated subsets have a large difference but the contaminated set has only two samples, that difference does not have a large impact on the model's evaluation score.
EPG plays a central role in our work, because we use it to empirically ground contamination metrics, assuming that metrics and corresponding thresholds that result in higher EPG give more adequate accounts of contamination. 

\subsection{Contamination threshold selection via z-score}\label{subsec:zscore}

As mentioned before, each contamination metric assigns a \emph{contamination score} to each sample in the evaluation corpus under scrutiny, after which a \emph{contamination threshold} is used to determine what threshold a model needs to exceed to be classified as contaminated.
At a contamination threshold of 0, any sample that has non-zero traces of contamination -- as defined by the metric -- is considered contaminated;
As we move the threshold up, samples get moved from the contaminated to the clean partition of the data, until a threshold of 1, when all samples are marked clean.

\begin{figure*}[ht]
    \centering
    \includegraphics[width=\textwidth]{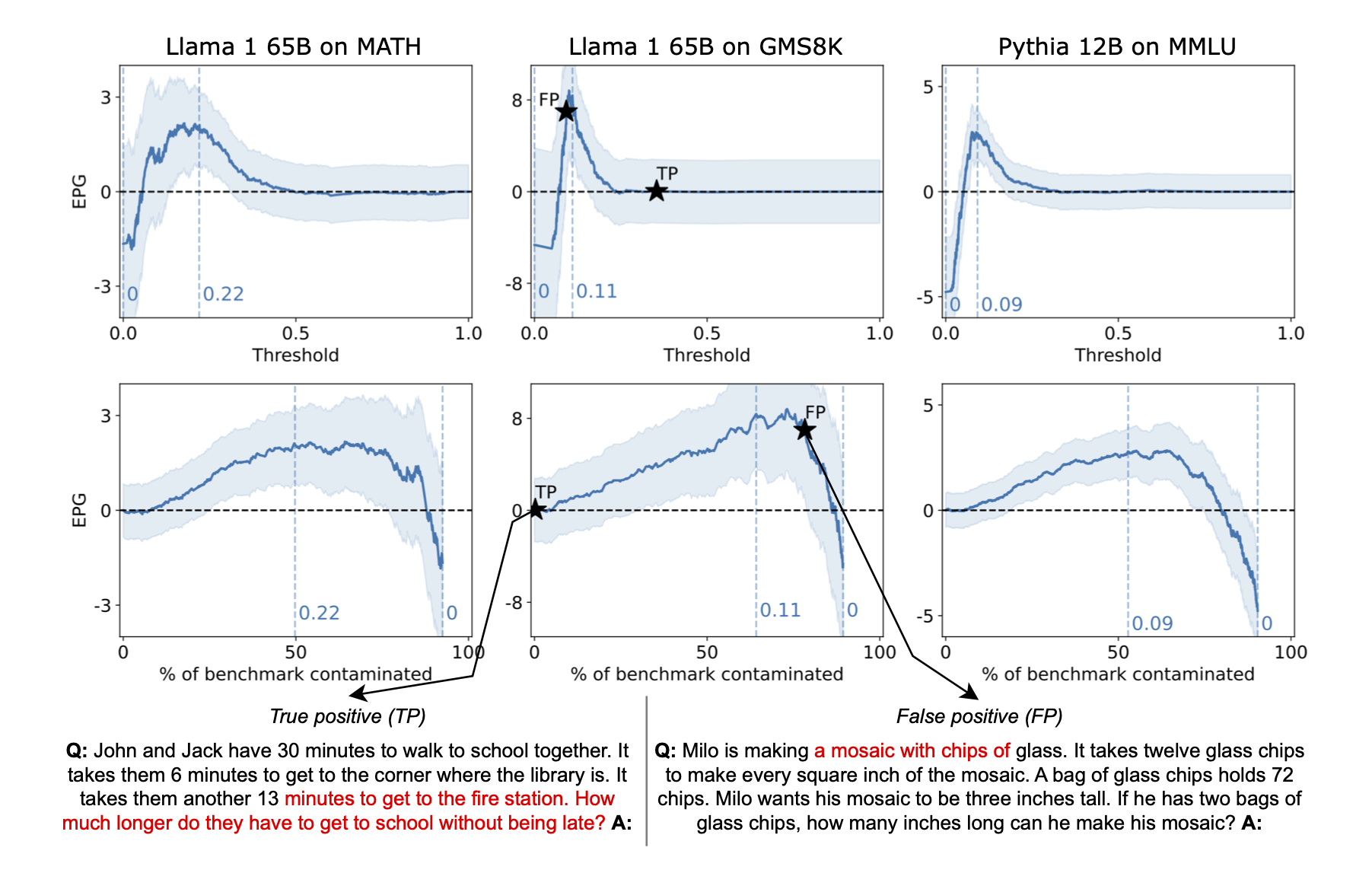}
    \vspace{-3em}
    \caption{\textbf{Benchmark analysis with threshold and \plotmethodname plots.}
    The threshold selected for a contamination metric has a large impact on the predicted amount of contamination.
    In the top row, we show how varying the threshold leads to different EPGs for a given model and benchmark pair. 
    At high thresholds, true positives may not be included, despite their effect on EPG; at very low thresholds, we instead see false positives that lower the EPG. 
    In the bottom row, we showcase our \plotmethodname plots, where the x-axis is instead the \% of data marked contaminated at a given threshold. 
    These plots offer a way of comparing different contamination metrics, as they solely analyse the ordering of data points enforced by a metric, rather than specific metric scores assigned to data points. 
    We show correspondence between points in the top row and bottom row and provide two examples of a true positive and a false positive for Llama 65B on GSM8k. 
    Optimal and zero thresholds are shown with vertical dotted lines.
    All lines correspond to scores from the \skipgram metric with $n=8$ and $skip\_budget=0$.}
    \label{fig:false_positives}
\end{figure*}

\paragraph{Importance of threshold selection}
The contamination threshold is an important knob to tune when considering and comparing contamination metrics.
However, it also makes it challenging to do direct comparisons of metrics across models, both because different models may respond differently to the same levels of contamination and because the same threshold has a different meaning for different underlying contamination metrics.
To address this issue, we select thresholds based on EPG.
Empirically, we have found it important to select thresholds separately for each model-benchmark pair, which we found to be crucial for eliminating false positives.
To explain why, in \cref{fig:false_positives} (top row), we show three examples of contamination analyses for different benchmarks, all of which exhibit false positives.
Each of these benchmarks contains samples with low, nonzero contamination score (toward the left of the plot), that do not lead to meaningful EPG.
As we move these examples to the `clean' partition of the dataset, EPG increases (as seen by the positive slope).
As a result, to meaningfully capture contamination, we have to select nonzero contamination thresholds to remove low scoring examples (false positives).\footnote{This addresses some of the issues discussed in \citet{yang2023rethinkingbenchmarkcontaminationlanguage}.} 
Notably, we find this method especially necessary for math world problems, where small snippets of context are often meaninglessly contaminated (e.g.\ ``a mosaic with chips of'').

\paragraph{Using z-scores}
Especially when the clean set is small, simply selecting the threshold that maximises EPG can sometimes be sensitive to noise.
To account for that, we consider a statistically-inspired approach where we estimate if performance on the clean subset is significantly different from the performance on the full benchmark. 
To quantify this, we compute the following z-statistic (per threshold):
$$
\text{z-score} = \frac{EPG}{err},
$$
where $err$ is the standard error for the given clean subset size (computed as $\sigma / \sqrt{N}$, where $\sigma$ is the standard deviation on the full benchmark).\footnote{We note some connection in this error metric and the 95\% CI's used by \citet{madaan2024quantifyingvarianceevaluationbenchmarks}.}
Selecting a threshold based on maximum z-score has the benefit of leading to a \textit{conservative estimate} of what percentage of a dataset is contaminated -- if two thresholds lead to the same EPG, the $err$ will be smaller for the one with a larger clean subset (and thus fewer samples marked contaminated) leading to a higher z-score.

\subsection{Qualitatively comparing contamination methods with \plotmethodname}
\label{subsec:qualitative_analysis}

Computing z-scores provides an easy method to compare contamination metrics across hyperparameters and thresholds.
However, finding the highest overall z-score implies taking a max in two different dimensions for each model, benchmark and pre-training corpus: first over the contamination threshold for a given contamination metric (incl.\ hyperparameters), and then over metrics (incl.\ hyperparameters). 
Even when optimising z-score rather than EPG directly, this method can yield insignificant results when comparing a large number of metrics across different hyperparameters.
Furthermore, it does not allow distinguishing metrics that perform similarly at a point estimate of a single threshold but have a different development across thresholds.
The max z-score may give the appearance that two metrics are close when they have a similar EPG at a single threshold value, while underneath the metrics lead to very different orderings of data points from ``most'' to ``least'' contaminated, in which the same samples get very different contamination scores.

\paragraph{\plotmethodname plots} To do a more in-depth comparison between methods, we therefore propose to directly consider how they order benchmark samples from least to most contaminated.
As the contamination threshold decreases, more and more points will be included in the contaminated set.
From this perspective, the effectiveness of the contamination scoring method can thus be interpreted in terms of how many true positives it adds as it removes false positives.%
\footnote{This can also be seen as minimising the number of false negatives (equivalent to maximising true positives) and false positives.}
To quantify this, we take inspiration from ROC-curves in classification and plot the EPG against the \% of the dataset marked as contaminated for different methods, allowing us to focus on the relative ordering of different points, which can be easily compared across methods.
We call such plots \textit{Contamination Threshold Analysis Method} (\plotmethodname) plots. 
Lines with steeper (positive) slopes correspond to finding more true positives for every false positive (lower false positive rate), while lines with negative slopes indicate the method is finding more false positives.
Lines that extend further to the right correspond to more data points having nonzero contamination score. 
If the EPG continues increasing, this would correspond to meaningful contamination (reduced false negatives), but if EPG decreases, this likely means false positives (like those of \cref{fig:false_positives}, bottom row) are being added.
In our analysis, we consider both z-score and \plotmethodname plots to better understand contamination metrics.

\section{Main results}\label{sec:results}

We compute contamination percentages and EPG for each benchmark, metric, hyperparameter set and threshold, for each of the seven models discussed in \cref{sec:methods:models} and their respective pre-training corpora.
In this section, we report two main results.
First, in \cref{subsec:comparing_methods}, we present a comparison across contamination metrics, providing both quantitative (\cref{subsec:numerical_results}) and qualitative results (\cref{subsec:qualitative_results}).
In this part of the results section, the main focus lies on the methods and whether they find meaningful EPG.
In \cref{subsec:found_contamination}, instead, we focus on how much contamination is found (\cref{subsec:how_contaminated}), what impact that has on various models' scores (\cref{subsec:how_impactful}), and on how that changes with model scale (\cref{subsec:scaling}).
After, in \cref{sec:analysis}, we dive deeper and investigate how various parameters impact the adequacy of contamination metrics.

\subsection{Detection methods across corpora, benchmarks and models}\label{subsec:comparing_methods}

Running all variations of all selected contamination metrics results in very large amounts of data; comparing that data across pre-training corpora and models directly is a non-trivial enterprise, especially when different thresholds and hyperparameters are involved.
As explained in \cref{sec:analysis_methods}, we rely on two main methods to compare different contamination metrics: we compute the maximum z-scores for each metric, per benchmark-model pair as well as across many of them, and we inspect how the EPG changes with the number of examples it marks contaminated.

\subsubsection{At a numerical glance}\label{subsec:numerical_results}
Before diving into a deeper and benchmark-specific comparison between metrics, we first present the average maximum z-scores for each detection method, across all seven model-pre-training corpus pairs.
In \cref{tab:comp_method_z}, we report both the averages as well as the number of model-corpus pairs for which significant EPG was found (column $N$).
In the following paragraphs, we discuss the main observations.

\begin{table*}[t]
\centering
\resizebox{0.85\textwidth}{!}{
\begin{tabular}{lcccccccc}
\toprule
& \multicolumn{2}{c}{\textsc{ngram-match}} & \multicolumn{2}{c}{\textsc{token-match}} & \multicolumn{2}{c}{\textsc{token-extend}} & \multicolumn{2}{c}{\textsc{longest-match}} \\
& \textit{avg} & \textit{N} & \textit{avg} & \textit{N} & \textit{avg} & \textit{N} & \textit{avg} & \textit{N}\\
\midrule
\textit{Big Bench Hard} & 9.2 & 7 & 9.1 & 7 & 10.6 & 7 & \textbf{12.1} & 7 \\
\textit{HumanEval} & 2.3 & 5 & \textbf{2.8} & 5 & 2.3 & 2 & 2.6 & 7 \\
\textit{HellaSwag} & \textbf{12.6} & 7 & \textbf{12.6} & 7 & 12.3 & 7 & 10.3 & 7 \\
\textit{MMLU} & 2.6 & 3 & 2.6 & 3 & 2.3 & 2 & \textbf{7.3} & 7 \\
\textit{PIQA} & \textbf{6.7} & 7 & 6.5 & 7 & 6.3 & 7 & 6.2 & 7 \\
\textit{GSM8K} & - & 0 & - & 0 & - & 0 & \textbf{3.4} & 4 \\
\textit{MBPP} & - & 0 & - & 0 & - & 0 & \textbf{2.3} & 4 \\
\textit{TriviaQA} & 4.5 & 6 & 5.1 & 5 & \textbf{5.4} & 7 & 4.9 & 7 \\
\textit{MATH} & - & 0 & - & 0 & - & 0 & \textbf{3.0} & 3 \\
\textit{Natural Questions} & 2.3 & 2 & 2.4 & 2 & \textbf{2.6} & 1 & \textbf{2.6} & 1 \\
\textit{DM Contest} & - & 0 & - & 0 & - & 0 & - & 0 \\
\textit{COPA} & - & 0 & - & 0 & - & 0 & - & 0 \\
\textit{SIQA} & - & 0 & - & 0 & - & 0 & - & 0 \\
\midrule
&6.8 & 37 & 6.9 & 36 & 7.7 & 33 & 6.3 & 54 \\
\bottomrule
\end{tabular}

}
    \caption{\textbf{Maximum z-scores for each metric.}
    We show the maximum z-scores for each metric and benchmark, across all models and pre-training corpus pairs where any significant EPG was found (\textit{avg})  and the number of model-benchmark pairs over which this average was computed (\textit{N}).
    The bottom row gives the per method weighted average z-score, and the total number of model-benchmark pairs for which significant EPG was found.
    All methods are run with their ``optimal'' hyperparameters ($n=8, mincount=1, skip\_budget=0$) -- see \cref{sec:analysis} for justification.
    }
    \label{tab:comp_method_z}
\end{table*}

\paragraph{No significant EPG for DM contest, COPA, and SIQA}
A first observation that can be made from \cref{tab:comp_method_z} is that no metric we considered found any meaningful contamination for DM Contest, COPA, and SIQA.
In Appendix \cref{fig:appx_comp_method} (top row), we can see that for the benchmark DM Contest, no method marked any data contaminated, suggesting that either none of the methods provides suitable signal for this model, or that the benchmark is truly entirely uncontaminated for the pre-training corpora at hand.
For the benchmarks COPA, and SIQA, on the other hand (rows 9 and 12 in the same figure), up to around 50\% contamination was detected, but there was no meaningful impact on performance for any of the models.

\paragraph{\skipgram provides overall best signal}
Secondly, we can see that by far the most significant EPG was found by the \skipgram method, which finds significant EPG for 10 out of 13 benchmarks, for 54 out of, in total, 70 of the model-corpus pairs across these 10 benchmarks.
Among all methods, it is the only method that finds significant EPG for the benchmarks GSM8K, MBPP, and MATH -- albeit not for all model-corpus pairs.
Furthermore, for six out of ten benchmarks, \skipgram finds more significant EPG (as quantified by average z-score across models) than the other methods.
The last row of the table, that provides the total number of model-corpus pairs for which a method finds meaningful EPG and the weighted average of the significance of the EPG it finds (as quantified by the z-score), also illustrates the danger of looking only at average numbers: \union clocks the highest average z-score, but only because it fails to detect meaningful EPG for three datasets for which \skipgram does find EPG -- but at lower levels.
In the next section, we further highlight such differences that cannot be observed directly by only looking at point estimates.

\captionsetup[subfigure]{justification=centering}
\begin{figure*}[t]
    \centering
    \begin{subfigure}[b]{\textwidth}
        \hfill\includegraphics[width=0.95\textwidth]{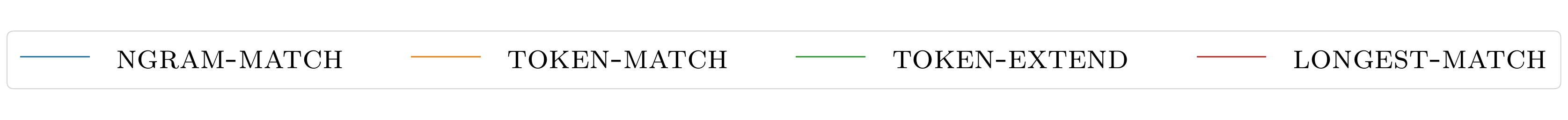}
        \vspace{-3mm}
    \end{subfigure}
    \begin{subfigure}[b]{0.26\textwidth}
        \includegraphics[height=3.87cm]{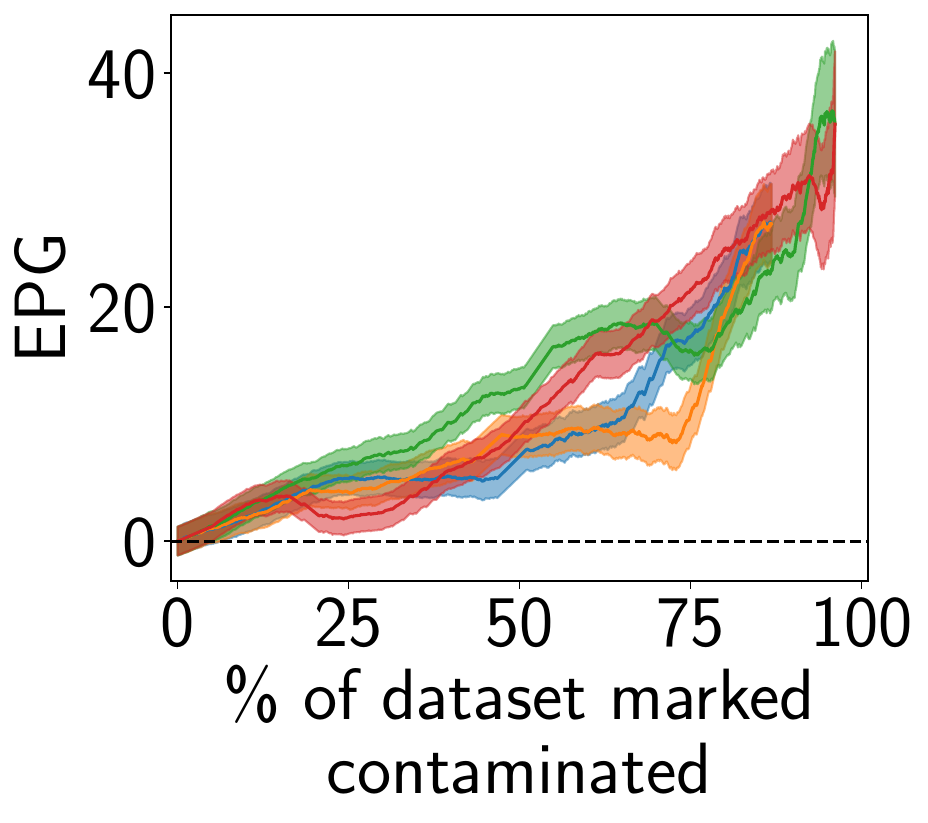}
        \caption{BBH\\ Llama 1 65B}\label{fig:main_comp_method_bbh_llama1}
    \end{subfigure}
    \begin{subfigure}[b]{0.24\textwidth}
        \includegraphics[height=3.87cm]{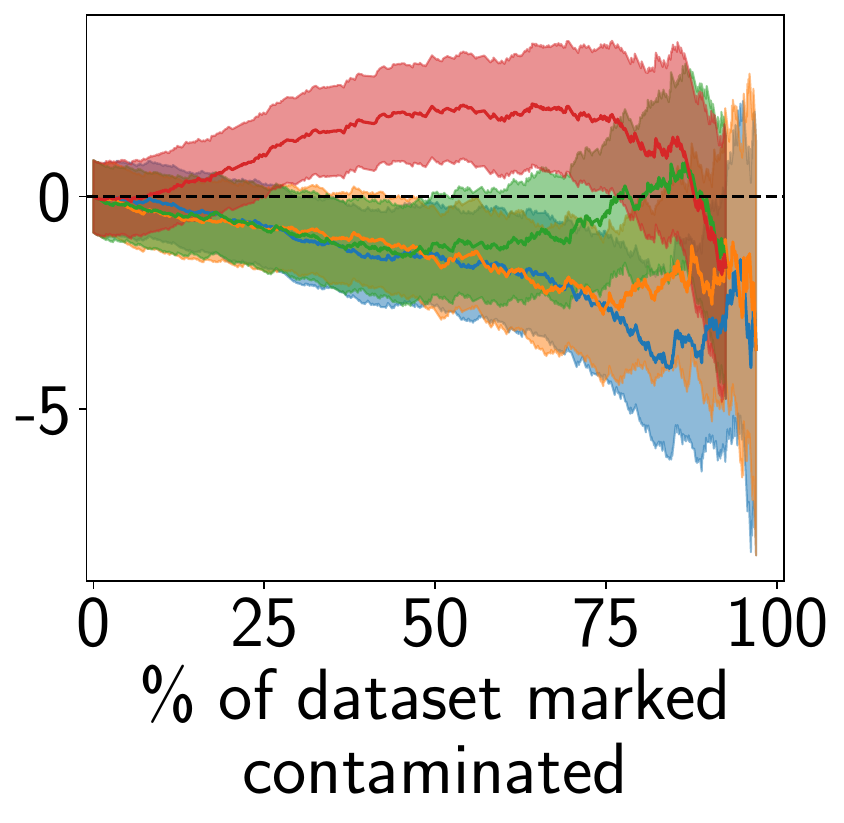}
        \caption{MATH\\ Llama 1 65B}\label{fig:main_comp_method_math_llama1}
    \end{subfigure}
    \begin{subfigure}[b]{0.24\textwidth}
        \includegraphics[height=3.87cm]{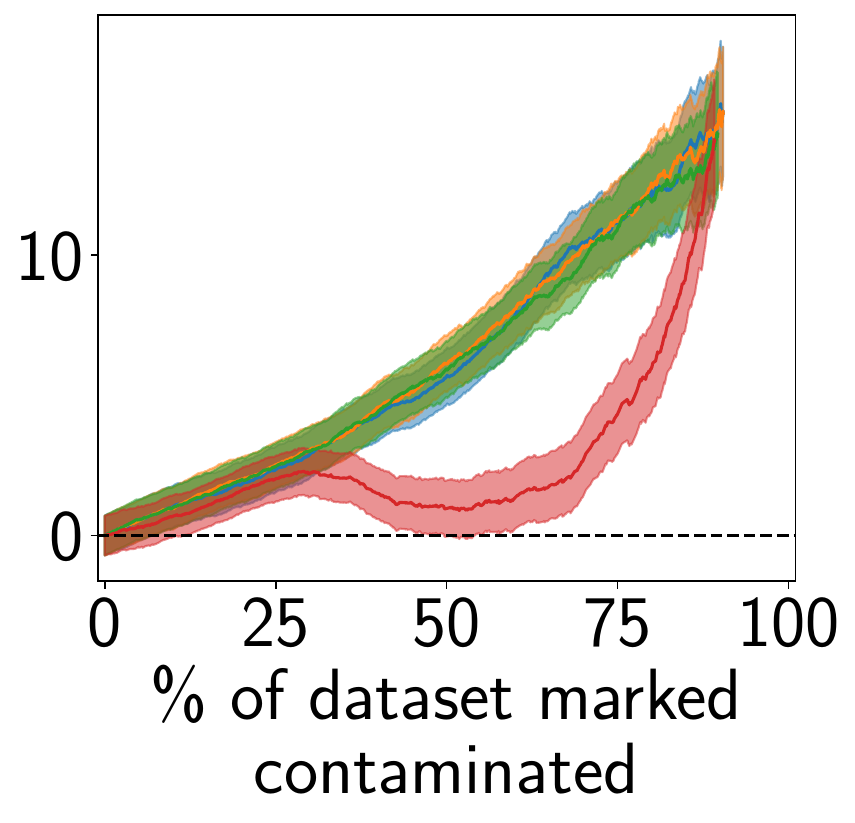}
        \caption{HellaSwag\\ Llama 1 65B}\label{fig:main_comp_method_hellaswag_llama1}
    \end{subfigure}
    \begin{subfigure}[b]{0.24\textwidth}
        \includegraphics[height=3.87cm]{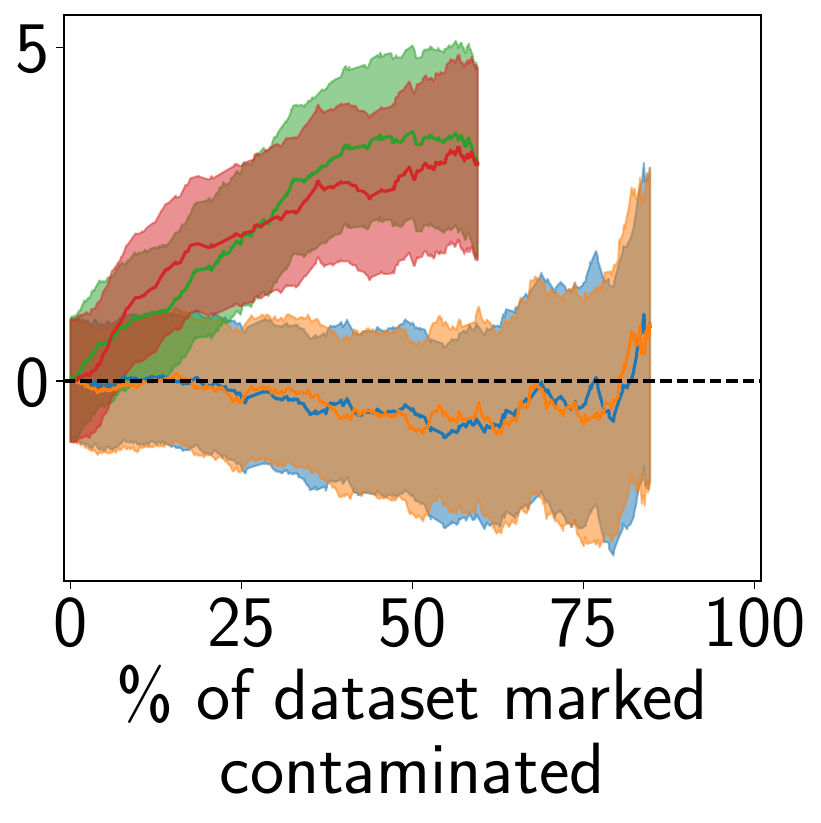}
        \caption{TriviaQA\\ Pythia 12b}\label{fig:main_comp_method_tqa_pythia}
    \end{subfigure}
    \caption{\textbf{Example profiles for different metrics.}
    (a) We most commonly observe cases in which all methods perform roughly the same or (b) \skipgram clearly performs best. 
    We also see more exceptional patterns: (c) \skipgram performs worse than other methods, except at threshold 0; and (d) \skipgram and \union perform equally well, outperforming \openai and \ngram. 
    Full results for all model-benchmark pairs can be found in \cref{fig:appx_comp_method}. 
    All methods depicted are run with their ``optimal'' hyperparameters ($n=8, mincount=1, skip\_budget=0$) -- see \cref{sec:analysis} for justification.
    }
    \label{fig:main_comp_method}
\end{figure*}

\subsubsection{Qualitative patterns across datasets and models}\label{subsec:qualitative_results}
Having reported overall numerical results for all methods, we now continue to discuss the qualitative patterns we observe across datasets and detection methods in our \plotmethodname plots, focusing still on only the best hyperparameters for each method.
Overall, we found that most model-benchmark pairs show one out of two common `profiles', with few exceptions.
We discuss those profiles as well as exceptions below, while the full set of \plotmethodname plots with curves for each metric (with optimal hyperparameters, see \cref{sec:analysis}), can be found in \cref{fig:appx_comp_method} for further inspection.

\paragraph{Pattern 1: \skipgram performs best}
A first prevalent pattern we observe is the case where \skipgram finds significant EPG where all other methods fail to.
An illustrative example is shown in \cref{fig:main_comp_method_math_llama1}.
As is clear in both \cref{fig:appx_comp_method} as well as \cref{tab:comp_method_z}, we observe this pattern for MBPP, HumanEval,%
\footnote{For Llama 1 65B, \skipgram is the only method that finds statistically significant significant EPG. At smaller sizes, the methods are similar.}
GSM8K, MATH, and MMLU. 
We find it especially interesting that \skipgram is specifically better at quantifying contamination on these tasks that are used to evaluate more sophisticated reasoning. 
Generally, the pattern is stronger for stronger models, with, for instance, significant EPG only being seen in Llama 1 models for GSM8K and MATH, despite contamination being detected in both the Llama pre-training corpus and the Pile.

\paragraph{Pattern 2: all methods are comparable}
A second common pattern that we observe is the case in which all methods perform roughly the same, with no significant differences in reported EPG per detected contamination.
For some benchmarks, such as Big Bench Hard (\cref{fig:main_comp_method_bbh_llama1}) and TriviaQA for Llama 1 (\cref{fig:appx_comp_method}, row 4), all metrics predict significant EPG, which increases as the contamination threshold relaxes and the percentage of the dataset marked contaminated increases.%
\footnote{TriviaQA is another interesting example where the average results in \cref{tab:comp_method_z} paint a somewhat misleading picture. Whereas it appears that \skipgram finds lower EPG than other methods, in \cref{fig:appx_comp_method}, we can see a slightly more nuanced picture: for the Llama 1 models, all methods give roughly the same result, whereas for Pythia, \skipgram and \union are similar but show a quite distinctly different pattern than \openai and \ngram.}
For other benchmarks, such as Natural Questions and to some extent COPA, the \plotmethodname curves trend positively, but our method does not allow us to statistically distinguish the results.

\paragraph{Pattern 3: Contamination for cloze tasks}
More incidentally, we observe a few other profiles, that are shared only across a few model-benchmark pairs.
One such profile is the one where \skipgram performs worse than other methods, except at threshold 0. 
We observe this behavior on HellaSwag (\cref{fig:main_comp_method_hellaswag_llama1}) and, to a lesser degree on PIQA, across models.
We hypothesise that this behaviour may be a result of the fact that both these benchmarks are evaluated in cloze format, by directly comparing the log likelihood of the possible answers.
Because of that, any occurrence of substrings occurring in the correct answer, even if unrelated, may contribute to increasing the likelihood of the correct answer string -- a hypothesis we explore further (with preliminary evidence) in \cref{fig:appx_comp_n_where}.
Especially at non-zero thresholds, the \skipgram method -- which counts only the longest match and thus discounts shorter matches stemming from different documents -- may thus miss subsequences that contribute to this increase in answer likelihood.
However, it is somewhat questionable whether the additional matches found by other methods at lower thresholds should really be counted as true positives, or are rather reflective of the limitations of cloze-style evaluation for LLMs.

\paragraph{Pattern 4: Normalisation makes a difference}
A last interesting pattern we observe is the one where \skipgram and \union perform equally well (within error bars) and better than \openai and \ngram (\cref{fig:main_comp_method_tqa_pythia}). 
We observe this only for Pythia models on TriviaQA, and hypothesise that it may be that normalisation -- not present in the former two methods -- may hurt contamination detection for this benchmark and pre-training corpus.

\subsection{How much contamination is found and how much does it matter}
\label{subsec:found_contamination}

The previous section illustrated that while \skipgram found overall the most meaningful EPG, there is no true `one-size-fits-all' method that clearly performs best across all datasets and models. 
Now, we consider how much contamination there is in the benchmarks we considered using the best metric and threshold for each (as reported in \cref{tab:optimal_thresh}) and how much impact that has on model performances (in the form of EPG).

\begin{table}[t]
    \centering
    \resizebox{0.95\textwidth}{!}{
    \begin{tabular}{lccccccccc}
\toprule
\multirow{2}{*}{Benchmark} & \multicolumn{5}{c}{Llama 1 Corpus} & \multicolumn{4}{c}{The Pile} \\
& Method & 7B & 13B & 33B & 65B & Method & 1.4B & 6.9B & 12B \\ \midrule
Big Bench Hard & \textsc{longest-match} & 0.05 & 0.05 & 0.06 & 0.05 & \textsc{ngram-match} & 0.04 & 0.06 & 0.06 \\
HumanEval & \textsc{longest-match} & 0.11 & 0.11 & 0.10 & 0.10 & \textsc{longest-match} & 0.10 & 0.11 & 0.12 \\
HellaSwag & \textsc{token-extend} & 0.26 & 0.26 & 0.24 & 0.24 & \textsc{token-match} & 0.20 & 0.17 & 0.17 \\
PIQA & \textsc{token-match} & 0.13 & 0.13 & 0.13 & 0.13 & \textsc{token-match} & 0.16 & 0.11 & 0.11 \\
MMLU & \textsc{longest-match} & 0.08 & 0.09 & 0.08 & 0.08 & \textsc{longest-match} & 0.12 & 0.12 & 0.09 \\
GSM8K & \textsc{longest-match} & 0.12 & 0.12 & 0.10 & 0.11 & \textsc{longest-match} & - & - & - \\
MBPP & \textsc{longest-match} & 0.08 & 0.08 & 0.08 & 0.08 & \textsc{longest-match} & - & - & - \\
TriviaQA & \textsc{longest-match} & 0.23 & 0.21 & 0.13 & 0.18 & \textsc{token-extend} & 0.35 & 0.33 & 0.34 \\
MATH & \textsc{longest-match} & 0.20 & - & 0.18 & 0.22 & \textsc{longest-match} & - & - & - \\
Natural Questions & \textsc{token-match} & 0.40 & - & - & - & \textsc{token-match} & - & - & 0.35 \\
\bottomrule
\end{tabular}
    }
    \caption{\textbf{Optimal contamination metrics per benchmark and pre-training corpus, with optimal model-dependent thresholds.}
    Dashes indicate benchmarks where no significant EPG was found for any metric or threshold. 
    We omit COPA, SIQA, and DM contest as no significant EPG was found for any model.
    In most cases, \ngram and \openai perform equally well (within error bars). 
    In those cases, we chose \ngram to simplify reporting. 
    We also provide optimal thresholds (based on maximising z-score in \plotmethodname plots) for each model scale (see \cref{subsec:scaling} for interpretation). 
    All methods are run with ``optimal'' hyperparameters ($n=8, mincount=1, skip\_budget=0$) -- see \cref{sec:analysis} for justification.}
\label{tab:optimal_thresh}
\end{table}

\begin{figure}[t]
    \centering
    \begin{subfigure}[b]{0.405\textwidth}
        \includegraphics[height=4.3cm]{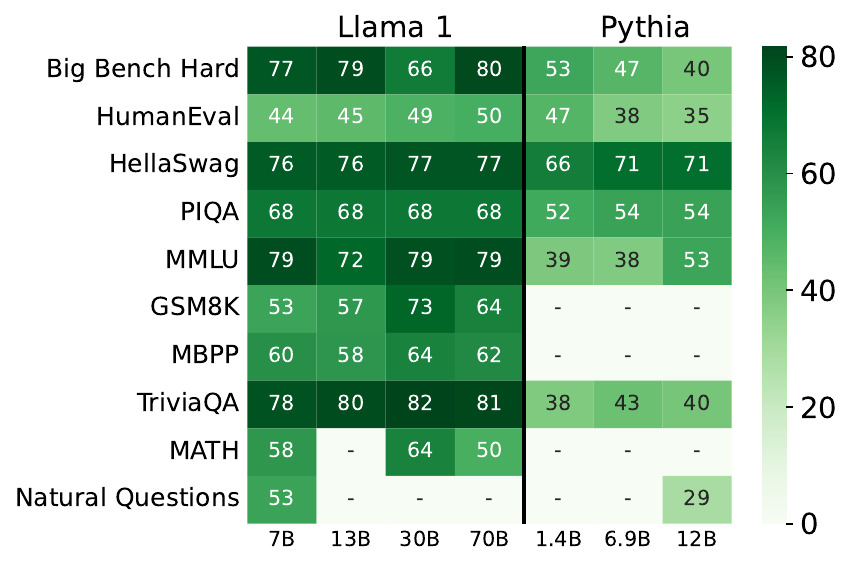}
        \caption{\% Contaminated}
        \label{fig:skip_pct_contam_all}
    \end{subfigure}
    \begin{subfigure}[b]{0.29\textwidth}
        \includegraphics[height=4.3cm, trim=43mm 0mm 0mm 0mm, clip]{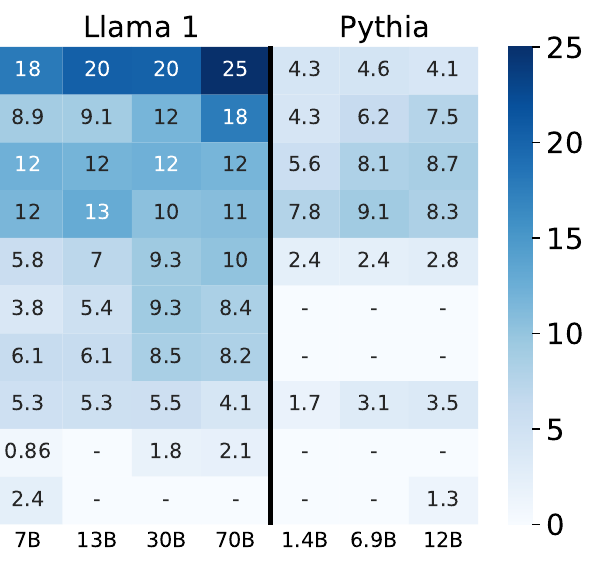}
    \caption{EPG}
    \label{fig:skip_epg_all}
    \end{subfigure}
    \begin{subfigure}[b]{0.285\textwidth}
        \includegraphics[height=4.3cm, trim=43mm 0mm 0mm 0mm, clip]{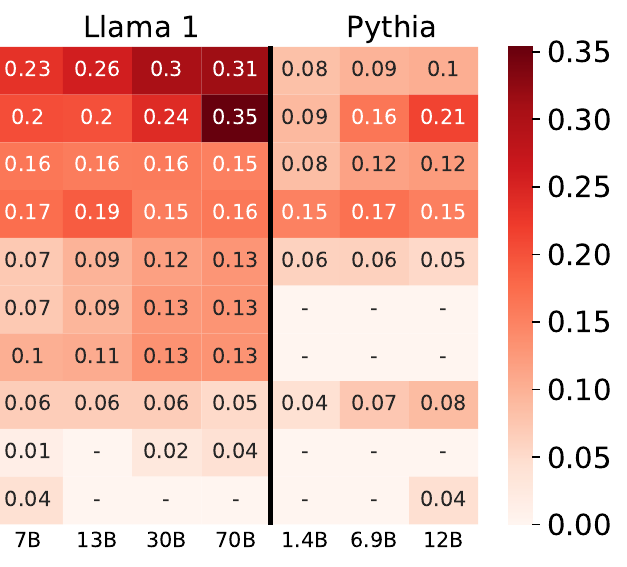}
    \caption{EPG per \% contam.}
    \label{fig:skip_epg_per_pct_contam_all}
    \end{subfigure}
    \caption{\textbf{Per model \% contaminated and EPG values across benchmarks.} Percentage of the dataset marked contaminated (a) and corresponding EPG (b) and average gain per \% contamination (c) for each of the model-benchmark pairs we considered in our study, according to the best contamination metric (see \cref{tab:optimal_thresh}). Optimal thresholds are selected separately for each model-benchmark pair. For COPA, DM Contest, and SiQA no significant EPG was found, and they are therefore omitted from the plot.}
    \label{fig:skip_epg_pct_contam_all}
\end{figure}

\subsubsection{How contaminated are benchmarks?}
\label{subsec:how_contaminated}
\begin{figure}[ht]
    \centering
    \begin{subfigure}[b]{0.397\textwidth}
        \centering
        \includegraphics[height=4.7cm]{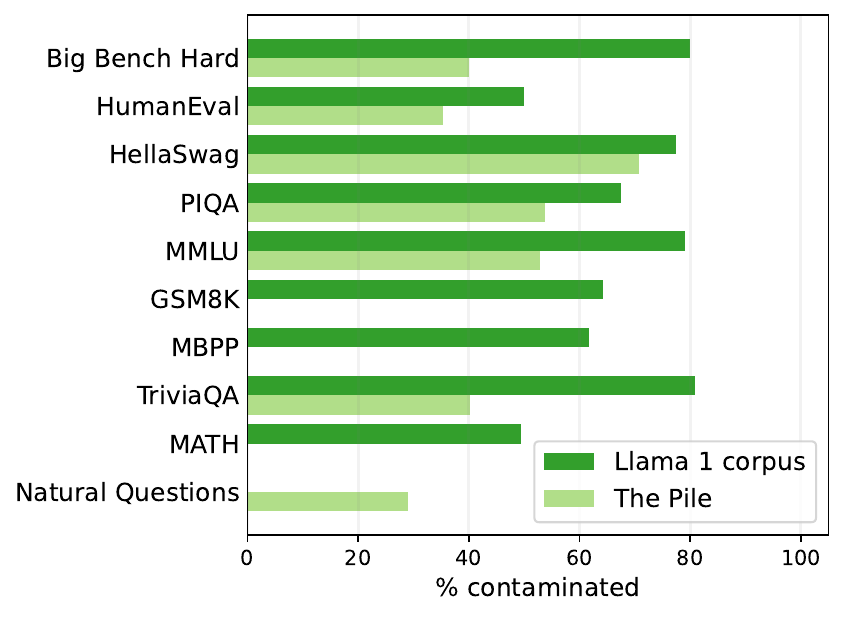}
        \caption{\% Contaminated}
        \label{fig:skip_pct_contam_best}
    \end{subfigure}
    \begin{subfigure}[b]{0.28\textwidth}
        \centering
        \includegraphics[height=4.7cm, trim=41mm 0mm 0mm 0mm, clip]{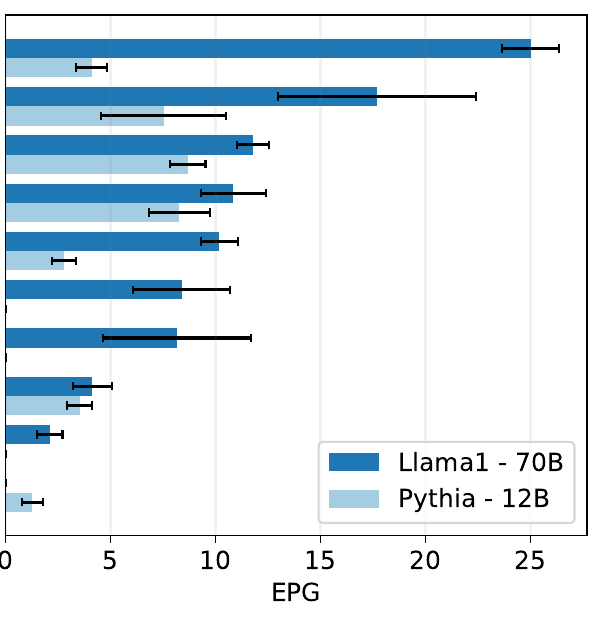}
    \caption{EPG}
    \label{fig:skip_epg_best}
    \end{subfigure}
    \begin{subfigure}[b]{0.28\textwidth}
        \centering
        \includegraphics[height=4.7cm, trim=41mm 0mm 0mm 0mm, clip]{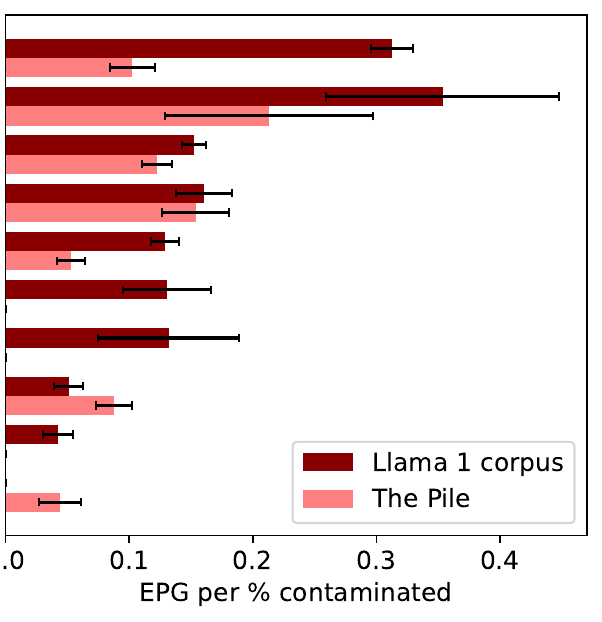}
    \caption{EPG per \% contaminated}
    \label{fig:skip_epg_per_pct_contam_best}
    \end{subfigure}
    \caption{\textbf{Percent contaminated, EPG and gain per \% contaminated for largest models.} We show the percent of each dataset marked contaminated (a), the corresponding EPG (b), and the gain per \% contaminated (c) for the largest two model sizes. With the exception of the benchmark Natural Questions, the Llama 1 corpus has substantially more contamination than the Pile, and also contamination scores are higher. Furthermore, with the exception of TriviaQA, the larger (and better) Llama models are also better able to exploit contamination, as indicated by the higher gain per \% contaminated plotted in (c).}
    \label{fig:skip_pct_contam_epg_best}
\end{figure}

In \cref{fig:skip_pct_contam_all}, we report the per-model values of the percent of each benchmark that is considered contaminated according to the best metric, hyperparameter set, and threshold, as reported in \cref{tab:optimal_thresh}. 
For easier comparison, we also plot the percent of contaminated data and resulting EPG for the largest two models of each family in \cref{fig:skip_pct_contam_best}, and the overall distributions in \cref{fig:skip_pct_contam_dist}.
For COPA, DM Contest, and SiQA no significant EPG was found, and we omitted them from these plots.

From \cref{fig:skip_pct_contam_all}, it is apparent that many of the benchmarks we considered are substantially contaminated in the Llama 1 pre-training corpus as well as in the Pile.
For 8 of the 13 datasets that we considered, on average more than 50\% of the samples are marked contaminated for the Llama 1 pre-training corpus.
Fewer contamination is found for the Pile, though also for that corpus the contamination numbers are substantially above zero for 6 out of 13 datasets, with percentages higher than or around 50\% for 4 datasets.
In particular Big Bench Hard, HumanEval, HellaSwag, MMLU, PiQA, and TriviaQA show substantial contamination levels across both corpora.
For MATH, GSM8K, and MBPP, no contamination resulting in EPG was found at all for the Pile, whereas contamination percentages in the Llama 1 corpus are substantial.%
\footnote{MATH provides another example where taking max and average numbers hides underlying patterns. 
As can be seen in \cref{fig:appx_comp_method}, row 6, \skipgram (red in the plot), provides a stable signal for Llama 33 and 65B. 
For Llama 7B and 13B, however, EPG only very slightly climbs when false positive examples are moved from the clean to contaminated set. 
For 13B, no single point estimate provides a significant z-score, while for 8B there is such a point estimate. 
Our z-score analysis, in this case, gives the impression that there is a vast difference between the 7B and 13B model.
In contrast, the plot illustrates that the effect of the detected contamination is in fact very similar.}

\subsubsection{Impact of contamination}
\label{subsec:how_impactful}
Having discussed what level of contamination is found, we now discuss the impact of that contamination and how that varies per model size and benchmark.
We show the across model EPGs in \cref{fig:skip_epg_all} and again show also results for the largest two models in \cref{fig:skip_epg_best}.
In both figures, we can see that not only the contamination percentages are larger in the Llama 1 corpus, but also the impact of the contamination is higher.
For the largest Llama model, both HumanEval and Big Bench Hard have an estimated increase in performance of more than 15\% (18\% and 25\%, respectively).
Three additional datasets (HellaSwag, MMLU, and PiQA) have an EPG of 10 points or higher.
For the smaller Pythia models, these numbers are substantially lower, ranging between 2 and 8 approximately.
A reasonable hypothesis for this difference could be that substantially more contamination was in fact found in the Llama 1 corpus.
However, as we can see in \cref{fig:skip_epg_per_pct_contam_all} as well as \cref{fig:skip_epg_per_pct_contam_best}, this offers only part of the explanation: the Llama models (with some exceptions) seem to generally be better at exploiting the contaminated examples, as illustrated by the higher gain they exhibit after normalising for the size of the contamination.
This difference may also explain the fact that for GSM8K, MBPP, and MATH, significant EPGs were found only in the LLama series of models (and not the Pythia series).
As can be seen in \cref{fig:appx_comp_method}, this is not because the corpus was free from matching $n$-grams: in the most lenient settings, up to around 90\% of the dataset could be marked contaminated.
However, while the contamination was exploited by the Llama 1 models, it did not result in any significant EPG for the Pythia models.
In upcoming sections, we will further discuss the relationship between model scale and ability to exploit contamination.

\subsubsection{EPG across model scale}
\label{subsec:scaling}

\begin{figure*}[t]
    \begin{subfigure}[b]{\textwidth}
    \centering
        \hspace{5mm}\includegraphics[width=0.37\textwidth]{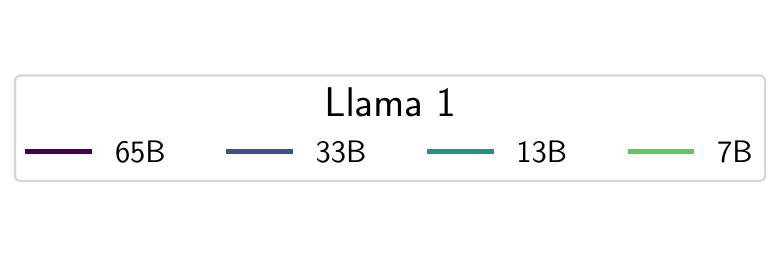}
        \vspace{-5mm}
    \end{subfigure}
    \begin{subfigure}[b]{0.345\textwidth}
        \includegraphics[height=4.5cm]{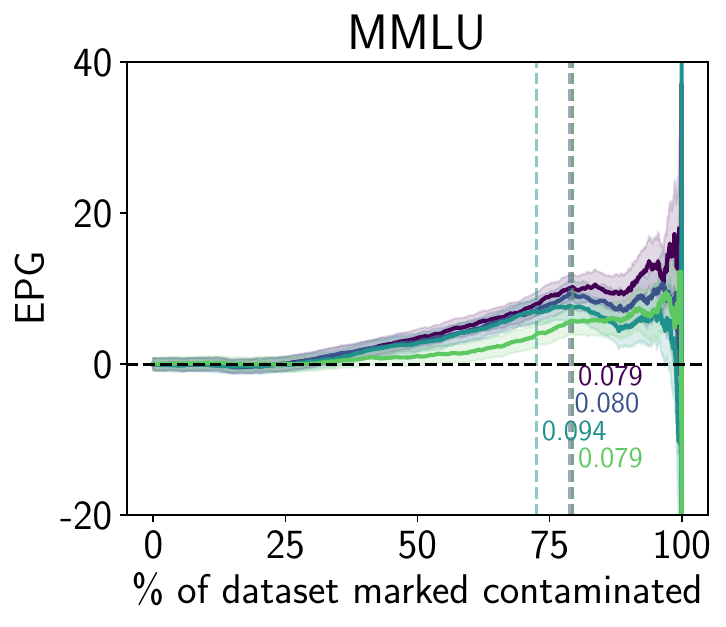}
        \vspace{-1mm}
        \caption{}\label{fig:scaling_llama1_gsm8k}
        \vspace{-2mm}
    \end{subfigure}
    \begin{subfigure}[b]{0.32\textwidth}
        \includegraphics[height=4.5cm]{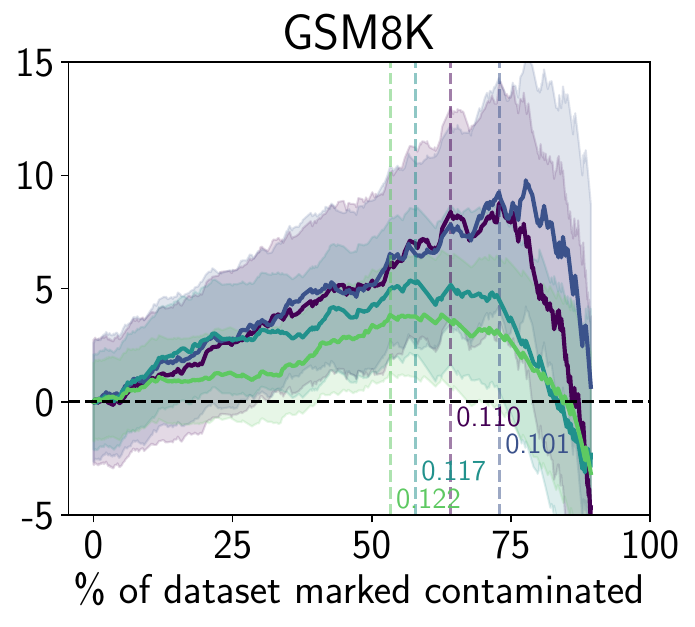}
        \vspace{-1mm}
        \caption{}\label{fig:scaling_llama1_mmlu}
        \vspace{-2mm}
    \end{subfigure}
    \begin{subfigure}[b]{0.32\textwidth}
        \includegraphics[height=4.5cm]{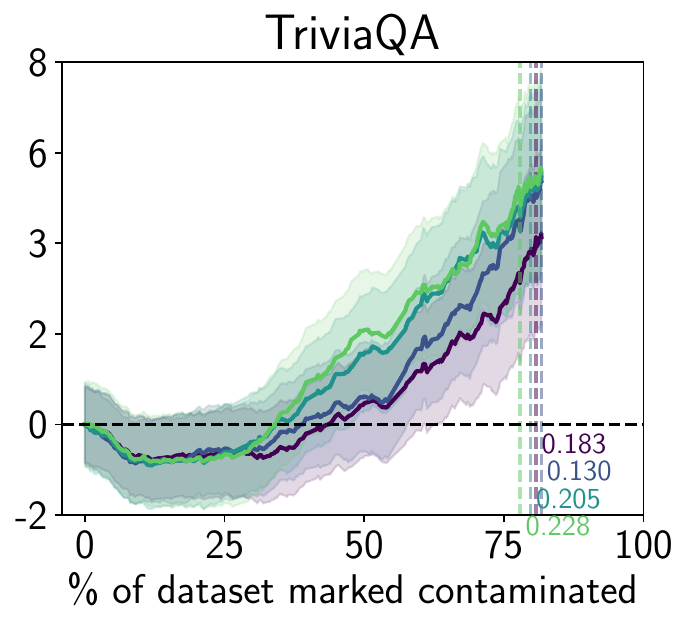}
        \vspace{-1mm}
        \caption{}\label{fig:scaling_llama1_tqa}
        \vspace{-1mm}
    \end{subfigure}
    \caption{\textbf{Example scaling behaviours for Llama models.} 
    Dotted lines indicate optimal contamination thresholds for each model size, chosen to maximize z-score.
    In (a), we see a shift primarily upwards, suggesting that larger models are better able to exploit contamination.
    In (b), instead, we see how contamination curves shift up and to the right as the model size grows, indicating that larger models benefit from examples that look like false positives for smaller models. 
    Lastly, in (c), we see the opposite case, where smaller models benefit more from contamination.
    We hypothesis that this is because larger models have high scores even on the clean partitions, and there is thus little room for improvement.
    }
    \label{fig:scale}
\end{figure*}

In \cref{fig:skip_epg_pct_contam_all} as well as \cref{fig:skip_pct_contam_epg_best}, we have seen that, for most benchmarks, the Llama 1 models profit more from contamination than the Pythia models, both in absolute and proportional terms.
Given that the Llama models generally also outperform the Pythia models (see \cref{fig:all_performances}), we hypothesise that the reason of this increased profit is that larger (or better) models may be better able to exploit contamination.
To qualitatively study the relationship between model size and ability to exploit contamination, we consider \plotmethodname plots for both model sizes in which we compare how EPG develops with the percent of the dataset marked contaminated by the \skipgram method, across model sizes.
In \cref{fig:scaling_llama_all} and \cref{fig:scaling_pythia_all}, we show results for all benchmarks, for both model series.
We see various different patterns, that we discuss next using a few representative examples shown in \cref{fig:scale}.

\paragraph{Cases where larger models exploit contamination more}
For around half of the model-benchmark pairs, there is a clear trend in which larger models appear to benefit more from contaminated examples. 
In \cref{fig:scaling_llama1_mmlu}, for example, we see that, for the Llama 1 series, the larger the model, the more it benefits from contamination, as exemplified by the positive y-shift in EPG scores for the same amount of contamination.
For Llama 1, we observe this pattern also for MATH, HumanEval, MMLU and -- to a lesser extent -- BIG-Bench Hard; for the Pythia models, we see it for HumanEval, Natural Questions, TriviaQA, MBPP, and HellaSwag.
In some cases, the optimal threshold does not change much (e.g.\ it hovers at $\approx$0.08 for MMLU, Llama 1), indicating that the same examples provide useful across model sizes, but larger models are simply better at exploiting them.
A perhaps more interesting case occurs when there is not only a positive shift on the y-axis, but also on the x-axis: for Llama 1, on some benchmarks, such as GSM8K (\cref{fig:scaling_llama1_gsm8k}) and HumanEval, the contamination curves shift up \textit{and to the right}, indicating that larger models can exploit examples that may look like false positives for smaller models.%
\footnote{An example could be the GSM8K question \emph{``Henry needs to assemble some toys, specifically 57 cars and 73 motorcycles. Henry knows that to assemble all the toys he will need 4 wheels for each car and 2 wheels for each motorcycle. How many wheels will be left if he has a box with 650 wheels in it?''} with a contamination score of 0.115. Llama 33B and 65B get this correct, but Llama 7B and 13B get it incorrect. The closest longest match in pre-training is "[84] cars and 73 motorcycles." (the brackets indicate a mismatch that stops the extension). Generally, we note it is difficult to ``conclude'' whether a contaminated substring is a true positive (given the inherent noise in language model evaluations) but we include this example as we find it unlikely that such strings are false positives, and performance on this data point indicates a true positive. Our primary approach for quantifying contamination relies on groupings of data points, avoiding these individual noisy estimates.}
For the Pythia models, we do not observe such an effect for any benchmark, and we hypothesise that their overall quality (see \cref{fig:all_performances}) may be too low to benefit in a way similar to the largest Llama models.

\paragraph{Cases where smaller models exploit contamination more}
A second pattern we observe is the exact opposite: for TriviaQA (\cref{fig:scaling_llama1_tqa}), HellaSwag, COPA, and PiQA, larger Llama 1 models benefit \emph{less} from contamination.%
We hypothesise that this may be because the larger Llama1 models perform very well on those benchmarks even without contamination, leaving less room for improvement exploiting contaminated examples.
\cref{fig:all_performances}, showing that the four previously mentioned benchmarks are indeed the highest performing ones for Llama 1, provides further evidence for this hypothesis.
Apart from Big Bench Hard, which appears to be an outlier, we do not observe this pattern for the Pythia models. 

\paragraph{No effect}
Lastly, for some benchmarks, there is virtually no effect.
In particular, for PiQA, HellaSwag, and Natural Questions, Llama 1 models have quite similar contamination curves; the same is true for GSM8K and PiQA.

\paragraph{Conclusion}
In sum, while there may be cases in which larger models can exploit contamination better, the extent to which contamination can be exploited -- evidently -- interacts with scale and/or how well a model can perform on a task by generalising from un-contaminated data.
Drawing overall conclusions around the impact of contamination across model sizes may therefore prove difficult, and we suggest that running separate threshold analyses for each model size is valuable to estimate the true impact of contamination.

\section{Analysis of parameter choices}
\label{sec:analysis}

In the previous section, we presented an overview of which metrics perform well and when, but without considering specific hyperparameter choices.
Here, we compare the impact of various parameters for the metrics we considered, such as the value of $skip\_budget$ where relevant (\cref{subsec:skip_budget}), the size of $n$ (\cref{sec:analysis_n}), and how frequent contamination is in the pre-training corpus (\cref{sec:analysis_frequency}).
In \cref{fig:appx_comp_n_where}, we also provide a brief analysis of the impact of the location of the contamination in the sample.

\subsection{The impact of $n$}
\label{sec:analysis_n}

\begin{figure}
    \input{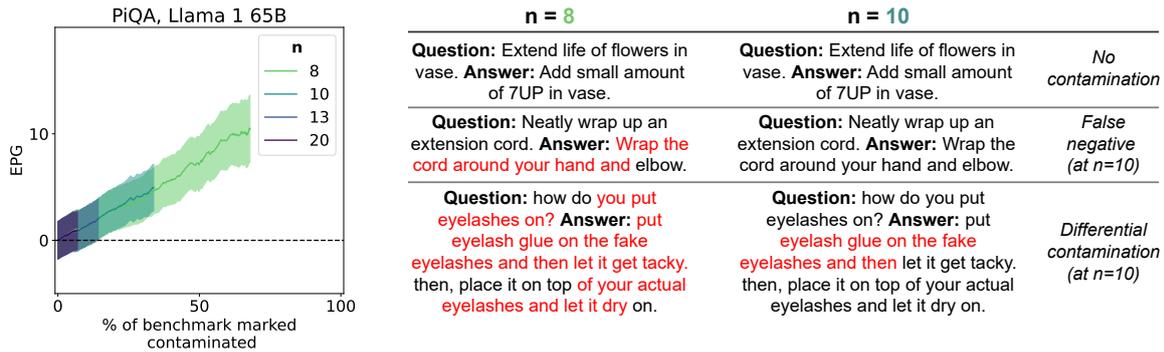}
    \vspace{-6mm}
    \caption{\textbf{Impact of $n$.} 
    On the left, we show a \plotmethodname plot for the \openai metric, for Llama 1 65 B on PiQA, for four values of $n$, with $mincount$ fixed at 1.
    This plot highlights the importance of using small $n$ for this \openai method, highlighting how at values of $n > 8$ many false negatives are seen. 
    For example, only 33.8\% of examples have nonzero contamination scores at $n=10$ while 67.9\% have nonzero contamination scores $n=8$. 
    These additional examples are classified as false negatives since they also correspond to EPG. 
    On the right, we see some example prompts from PiQA, with contaminated substrings highlighted in red. 
    These example illustrate how a simple choice, such as changing $n$ can have varied effects on contamination scores (top row: no change, middle row: contamination where there seemed to be none, bottom row: significantly more contamination than there may have seemed).}
    \label{fig:piqa_false_neg}
\end{figure}

First, we consider one of the most important parameters in $n$-gram based overlap methods: the value of $n$ used in the experiments.
Prior work has simply chosen a value for this hyperparameter.
Here, we analyse its effect by considering $n\in\{8, 10, 13, 20\}$.
Specifically, we use \plotmethodname to understand how the value of $n$ impacts the resulting EPG.
We show the results in \cref{fig:appx_comp_n} and \cref{fig:appx_comp_n_skipgram} for all model-benchmark combinations for the \openai and \skipgram metrics, respectively.
We find that, almost across the board, values of $n$ larger than $8$ lead to false negatives (examples that result in increased EPG but are not identified as contaminated for higher values of $n$).
To exemplify this, we provide a worked-out example in \cref{fig:piqa_false_neg}, for the \openai metric. 
For $n\in\{10, 13, 20\}$, the number of contaminated examples never exceeds 10\%, 20\%, and 40\%, respectively, even at the lowest threshold, and the maximum EPG that is found for Llama 1 65B is around 4.
As we move to $n=8$, we can see that both the percent of the benchmark it marks contaminated and the resulting EPG increase, indicating that many examples are found that were labeled clean when using n=$10$ or larger, but that do result in increased EPGs.
We find similar trends across most benchmarks and models.


\subsection{Impact of the skip budget}
\label{subsec:skip_budget}

\begin{figure}[t]
\centering
    \begin{subfigure}[b]{0.5\textwidth}
        \centering
    \resizebox{0.75\textwidth}{!}{
        \begin{tabular}{lcc}
\toprule
 & Llama 1 & The Pile \\
 \midrule
DM Contest & 0.987 & 0.988 \\
Big Bench Hard & 0.995 & 0.997 \\
Natural Questions & 0.986 & 0.997 \\
TriviaQA & 0.978 & 0.989 \\
GSM8K & 0.985 & 0.992 \\
MATH & 0.979 & 0.985 \\
HumanEval & 0.986 & 0.987 \\
MBPP & 0.994 & 0.997 \\
COPA & 0.992 & 0.997 \\
HellaSwag & 0.989 & 0.995 \\
PIQA & 0.996 & 0.997 \\
SIQA & 0.994 & 0.997 \\
MMLU & 0.991 & 0.993 \\
\bottomrule
\end{tabular}
    }
    \caption{}
    \label{tab:spearman_skip}
    \end{subfigure}
    \hspace{-15pt}
    \begin{subfigure}[b]{0.5\textwidth}
        \centering
        \includegraphics[height=6.7cm]{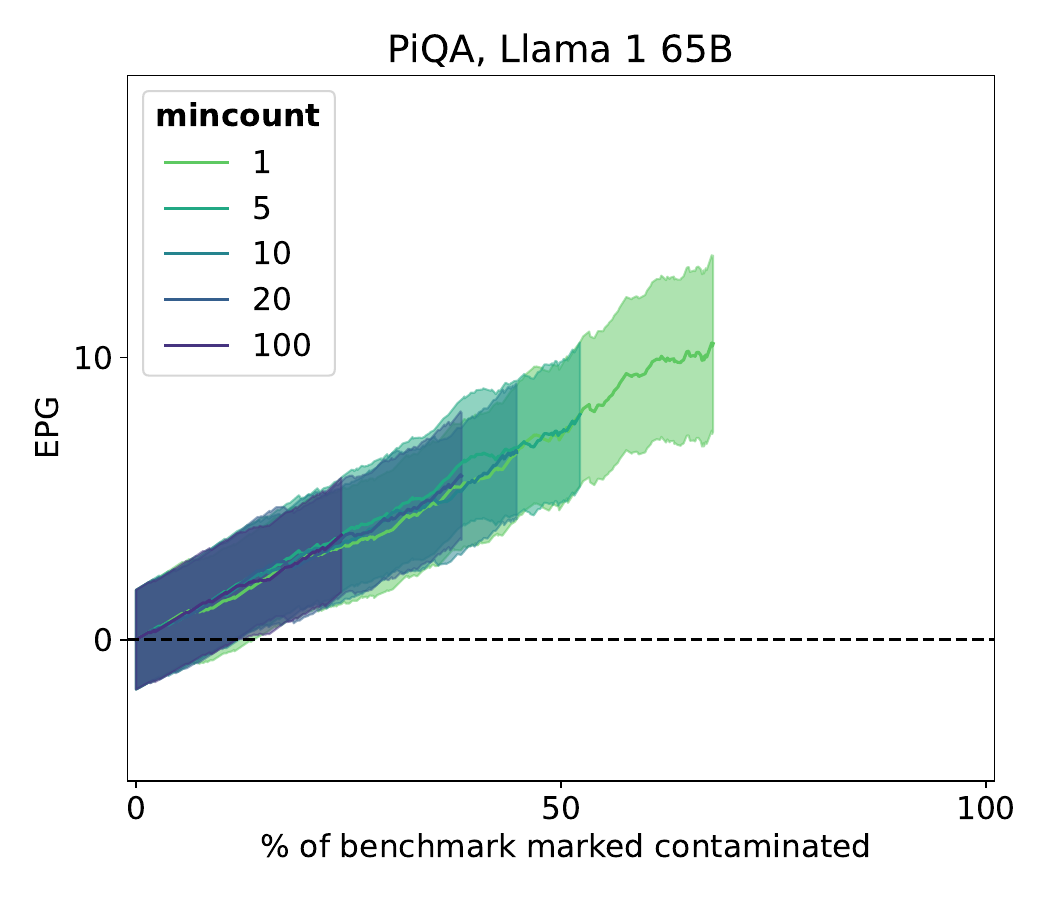}
        \vspace{-3mm}
        \caption{}\label{fig:impact_mincount}
    \end{subfigure}
    \caption{\textbf{Spearman rank correlations for \union and impact of $mincount$.} (a) Spearman rank correlations ($\rho$) between the ordering of examples for each benchmark and pre-training corpus when using \union with $skip\_budget = 5$ vs $skip\_budget = 0$. (b) \plotmethodname plot showing the impact of $mincount$ for PiQA, Llama 1 65B for the \openai metric.}
    \label{fig}
\end{figure}

Both the \skipgram and \union methods rely on a \emph{skip budget}, which allows for minor differences within the matched substrings.
Here, we investigate the impact of the size of this skip budget on the results.
Surprisingly, we find that specific choice of $skip\_budget$ to make little-to-no difference (\cref{fig:appx_comp_skip}). 
While the precise scores assigned to individual examples do change -- as one would expect, the scores are higher for higher values of $skip\_budget$, see \cref{tab:appx_comp_skip} -- the overall effect on EPG seems largely the same. 
This may seem counterintuitive, but makes sense considering that the overall ordering of points remains largely the same. 
We quantify this intuition in \cref{tab:spearman_skip} by computing the Spearman rank correlation ($\rho$) between the ordering of data points when using contamination scores with $skip\_budget = 0$ and $5$, finding that all these correlations are extremely high.
We suspect this is due to the fact that $skip\_budget$ only accounts for substitutions -- ideally, a more sophisticated matching algorithm would have a budget for insertions or deletions, and thus better be able to account for inserted content in evaluation samples (as compared to their possible ``raw'' contaminations in pre-training data).

\subsection{The frequency of contamination in the pre-training corpus}
\label{sec:analysis_frequency}

In computing contamination scores, we generally do not consider \emph{how often} a particular match occurs in the pre-training corpus.
Yet, it is reasonable to assume that if an example (partly) occurs once in a very large corpus, that is more difficult to exploit than if it occurs multiple times.
To investigate this hypothesis, we compare the \openai method across various $mincount$s, that specify how often a match should minimally occur in the training corpus to count.%
\footnote{Note that $mincount$ is hard to define for \union and \skipgram. 
If an extension of twice the length appears once, but the seed string of half the length appears 10 times, which one should we use in generating a contamination score? 
We leave such investigations to future work.}
Somewhat surprisingly, we find that, similarly to the finding of using small $n$, setting $mincount$ to values higher than 1 results may exclude examples that do have a real increase in performance as a consequence (see \cref{fig:impact_mincount} for a representative example).
We generally find similar behaviours (see \cref{fig:appx_comp_mincount} for a plot containing the full sweeps) to the case of decreasing $n$, indicating that higher $mincount$ values also simply lead to more false negatives (or have no effect). 
Overall, these two findings point to the fact that prior works using the \openai method \citep{brown2020fewshot,chowdhery2022palm} are likely too strict in what they mark as contamination, which may explain why they seemed to find that contamination has little effect.%

\section{Conclusion}
\label{sec:conclusion}

In this paper, we presented an elaborate investigation into evaluation data contamination, with a special focus on how the choice of contamination metrics and its hyperparameters impact conclusions about contamination levels and their impact.
Specifically, we conduct a study in which we compare four contamination metrics -- three from the literature and one new -- that define contamination in terms of overlapping $n$-gram spans (termed ``matches'') between a sample in the evaluation benchmark and the pre-training corpus.
We analyse these metrics across \emph{5 parameter settings}, on \emph{13 benchmarks}, for \emph{7 models} of various sizes trained from \emph{2 pre-training corpora}.

\subsection{Methodology} 
To quantitatively compare metrics, we consider the \emph{estimated performance gain} (EPG) that they predict from the detected contamination, operating under the assumption that a useful contamination metric flags contamination that has an actual impact on model performance.
In our comparisons, we rely on a quantification of the statistical difference between score distributions on the full benchmark and the subset marked by the various metrics as non-contaminated as measured with z-scores, supplemented by a novel method we call the \textit{Contamination Threshold Analysis Method} (\plotmethodname).
Where the former allows for a numerical comparison between metrics and/or hyperparameters, the latter provides a qualitative insight in how different contamination metrics and hyperparameter order samples from least to most contaminated, illustrating differences that a pure numeric approach would not reveal.

\subsection{Summary of results}
Using these two methods, we conduct extensive experimentation across metrics and corresponding hyperparameter choices.
In our investigation, we first consider if there is any best metric, across benchmarks and models, which we discuss in \cref{subsec:comparing_methods}.
In this section, we discuss different patterns across benchmarks and models.
Our main conclusions are that our novel \skipgram metric, which considers only the longest overlapping substring rather than a union of all of them, provides a stable signal across the board.
In some cases, its results are comparable with the other metrics with best hyperparameter settings; in others, the metric is able to find contamination that no other metric could.
We also show how good threshold selection can help minimise the false positives and negatives given a fixed contamination metric.

Then, in \cref{subsec:found_contamination}, we report contamination results for the 13 benchmarks we consider, for seven different models consisting of four Llama 1 and three Pythia models.
For each model, benchmark pair, we report results for the metric that finds the most meaningful EPG according to our analysis metrics, with optimal hyperparameters (which metric that is and with which thresholds can be found in \cref{tab:optimal_thresh}).
One of our main findings in that section is that the impact of evaluation data contamination has been underestimated in many prominent LLM releases (Figure \cref{fig:skip_epg_pct_contam_all,fig:skip_pct_contam_epg_best,fig:skip_epg_pct_contam_dist}, among others).
While previous work often reported substantial amounts of contamination (see \cref{tab:lit_overview_table}, column 5), they also note that the impact of this contamination on benchmark scores is negligible or even negative (\cref{tab:lit_overview_table}, column 6).
Our work shows that these results are likely driven by too many false positives or negatives in the detected contamination, resulting from suboptimal (parameter selection in the chosen) contamination metrics.

Lastly, in \cref{sec:analysis}, we investigate the effect of various metric parameters on the detected levels of contamination and their impact.
We find that using smaller values of $n$ gives better results across the board (\cref{sec:analysis_n}): using values of $n$ larger than 8 increased the number of false negatives for virtually all benchmarks (and across metrics).
Somewhat surprisingly, we find that the same is true for the $mincount$ parameter that specifies what the minimal occurrence of an overlapping string in the pre-training data should be for it to count as contaminated (\cref{sec:analysis_frequency}).
While it stands to reason to think that the frequency of an overlapping string has an impact on the extent to which it influences benchmark scores, in all experiments we did, discounting overlapping strings that occurred more than once increased the number of false negatives more than it reduced the number of false positives.
Lastly, in \cref{subsec:skip_budget}, we studied the impact of the skip budget in the \union metric and found that it has little impact.

\subsection{Recommendations for contamination analysis}
Our investigation gives a nuanced account of contamination and its impact, showing how the impact of contamination differs across models and sizes and that something that is beneficial for one model may not be for another.
At the same time, the results do paint a consistent picture: the `best' method is consistent across model families and sizes, the optimal threshold is substantially more stable within than across benchmarks (std 0.02 within benchmark vs 0.1 across benchmark means), and there are virtually no cases in which the overall best method is substantially worse any of the other methods we considered.
These consistencies make it thus easy to provide concrete suggestions for doing contamination analysis:\begin{enumerate}\setlength\itemsep{0.1em}
\item Running multiple contamination metrics is preferred, but if you only run one, use \skipgram;
\item Use z-scores supplemented by \plotmethodname to do benchmark specific threshold selection. This is cheap, as it can be done post-hoc;
\item Report \% contaminated, EPG and the selected thresholds. If space allows, show also \plotmethodname plots for further context.
\end{enumerate}

\noindent If possible, supplementing this with manual analysis to check which samples are flagged is preferred.
We hope our work encourages more consistent reporting of the effects of contamination on LLMs.

\section{Discussion}
\label{sec:discussion}

In this work, we proposed to compare contamination metrics post-hoc through the effect that the samples they detect have on benchmark performance.
This proposal is based on several assumptions.
In this last section, we discuss those assumptions and their limitations in more detail.

\subsection{No effect =? no contamination}
One of the most central assumptions underpinning our methodology is that useful contamination metrics detect contamination that has an actual impact on performance.%
While this standpoint is very defendable from an evaluation perspective -- if contamination has no impact on model performance, it arguably does not matter much for how we interpret benchmark scores -- it also has some clear limitations.
Specifically, there is a range of conclusions that cannot be drawn using our methodology.

First, \plotmethodname does not allow to detect contamination if too little or too much of the benchmark is marked as contaminated.
Both when the clean set is too small or the contaminated set is too small, it is not possible to get a statistically significant signal on the difference between the full benchmark and the clean partition.
This implies that if almost all of a benchmark is marked contaminated, our method cannot be used to determine whether those samples are true or false positives.
Additional analysis, such as manual inspection of the samples or perhaps approaches that directly address memorisation (see \cref{subsec:related_memorisation}) would be needed in such a case.

Second, because it directly links contamination to its impact, our method does not directly allow for the (possible) conclusion that contamination has no impact on model scores.
In our work, there were some cases where large percentages of contamination without any EPG, such as for the benchmarks MATH, GSM8K and MBPP for the Pythia models.
With our methodology, those examples are considered to be false positives.
It is, however, also possible that the Pythia models are simply not able to benefit from the contaminated examples, perhaps because they lack the capacity to memorise the examples.
While this may not matter from a practical perspective, it is scientifically interesting to consider the impact of contamination and levels of contamination separately.
Doing so would require other methods than the one proposed in this paper.

Lastly, it is important to note that our method is only suitable to compare metrics that are a priori likely to capture forms of contamination.
Doing a gridsearch through metrics that are not may result in preferring metrics that flag examples that are \emph{easy} and therefore have better scores.\footnote{Though it is relatively unlikely that such a protocol would lead to a metric that generalises across benchmarks.}
On the other hand of the spectrum, metrics that are too lenient may detect pre-training data that is beneficial for a model because they facilicate generalisation.
Any trained model that performs better on a benchmark than a random model must have necessarily inferred its ability from somewhere in the training corpus, and the pre-training data that it used to do so is not always to be considered contamination.
We will further discuss this in the next subsection.

\subsection{Text-based vs embedding based}
In our work, we considered only text-based contamination metrics.
Such metrics are limited in the type of contamination that they can detect. 
Specifically, they cannot detect types of contamination that consist of paraphrases of benchmark examples, and they are sensitive to minor perturbations in the prompt.%
\footnote{The skip budget accounts for this, to some extent, but is nevertheless limited because it allows for substitutions only, not insertions, and was shown to have little impact in our experiments.}
In our piloting stage, we conducted a small-scale experiment in which we used similarity search \citep{douze2024faisslibrary} to detect contaminated examples.
In this pilot, we embedded a subset of the training corpus and used product quantisation \citep{jegou2011product} to partition the training corpus embeddings according to the subvector centroids of the quantizer, analogously to the $n$-gram matching approach, in order to bypass indexing the embeddings.
Our experiments showed that using similarity search is feasible to detect similar samples. 
However, the initial approach suffered from recall issues, and further studies are required to determined the optimal set up.
Furthermore, using embedding-based approaches may make it more challenging to distinguish examples that contribute to generalisation from examples that contribute to memorisation.
Consider, for instance, the case of factual knowledge tasks: to perform well on such tasks, the knowledge needs to be in the training data in some form or another.
While a verbatim copy of the evaluation question should be considered contamination, other semantically similar phrases should likely not be and are valuable for knowledge intensive tasks..
Nevertheless, we believe that comparing embedding- and text-based approaches and their differences is an interesting direction to explore in future work.
Especially supplementing the two approaches could lead to new scientific insights, not only related to evaluation data contamination, but also more broadly to the relationship between pre-training data and model outputs.

\subsection{Post-hoc vs causal contamination}
Lastly, in our work, we adopted a post-hoc approach to contamination detection.
All our analyses are thus fundamentally correlational.
An interesting experiment to supplement our analysis and judge the extent to which it is obfuscated by confounds would be to do a causal intervention to study if removing the contaminated examples does in fact lead to a comparably lower score.
However, causal contamination experiments are expensive, as they require re-training a model.
Especially given our results in \cref{subsec:scaling}, which indicate that contamination effects may vary which scale, such experiments would likely have to be conducted at scale, further increasing their cost.
While some work has considered finetuning on benchmarks to get some notion of causal effects \citep[e.g.][]{jiang2024investigatingdatacontaminationpretraining}, the connection to at-scale pre-training is still missing.

\subsection{Pre- vs post-trained models}
In our work, we focused only on pre-trained models.
Post-training methods such as SFT and RLHF as well as test-time methods that work on model inference are known to change model behaviour, which can potentially also impact how models can benefit from knowledge in the pre-training corpus.
We consider a study into the effects of contamination in post-trained models as an interesting direction for future work.


\section{Acknowledgements}
We thank Lovish Madaan for help with debugging experiments.
We thank Wojciech Galuba for the ground work on setting up the code base for running text-based overlap metrics before this project started.
We thank Andrew Saxe for his support on the project.

\bibliography{custom,anthology}
\bibliographystyle{acl_natbib}

\newpage

\appendix

\section{Overview of previously used contamination metrics}\label{sec:lit_overview}

In \cref{tab:lit_overview_table}, we provide an overview of the various contamination metrics used in previous work, the benchmarks that the respective papers computed these metrics for, and -- as much as possible -- the approximate results achieved with these metrics.
Whenever possible, we reported the impact in terms of EPG.
In some cases, the information to do so was not sufficient, or particular metrics were not provided. These are marked with NA or presented in the format they were shared in the paper.

\scriptsize
\scriptsize
\begin{longtable}{lcllll}
        \toprule
        \emph{Model} & \emph{Overlap} & \emph{T} & \emph{datasets} & \emph{\%contam.} & \emph{impact} \\
        \midrule
        \makecell[l]{PALM 8B \\ \citep{chowdhery2022palm}}
        & 13 tokens & 0.7 & \makecell[l]{TriviaQA, Lambada,  WebQuestions, WSC, Winograd,\\ SQuADv2, ARC-e, ARC-c, CB, ReCoRD} & 85-20\% & -0.6-4.4 \\
        \midrule
        \multirow{3}{*}{\makecell[l]{Llama 2 7B\\ \citep{touvron2023llama2}}}
        & \makecell[l]{10 tokens, 4 skips} & 0.8 & \makecell[l]{HellaSwag, MMLU, MMLU-hum} & 8-11\% & 1-1.8 \\
        & \makecell[l]{10 tokens, 4 skips} & 0.2 & \makecell[l]{HellaSwag, MMLU-Hum} & 15-26\% & 2.8-2.1 \\
        & \makecell[l]{\{10,20,\ldots,50\} tokens\\ 4 skips} & 0.2 & \makecell[l]{TriviaQA, NQ, BBH, GSM8K, HumanEval} & NA & 0 \\
        \midrule
        \multirow{3}{*}{\makecell[l]{Llama 2 70B\\ \citep{touvron2023llama2}}}
        & \makecell[l]{10 tokens, 4 skips} & 0.8 & \makecell[l]{HellaSwag, MMLU, MMLU-hum} & 8-11\% & 0.9-1.2 \\
        & \makecell[l]{10 tokens, 4 skips} & 0.2 & \makecell[l]{HellaSwag, MMLU-Hum} & 15-26\% & 0.9-3.1 \\
        & \makecell[l]{\{10,20,\ldots,50\}\\tokens, 4 skips} & 0.2 & \makecell[l]{TriviaQA, NQ, BBH, GSM8K, HumanEval} & NA & 0 \\
        \midrule
        \makecell[l]{PALM 405B \\ \citep{chowdhery2022palm}}
        & 13 tokens & 0.7 & \makecell[l]{TriviaQA, Lambada, WebQuestions, WSC, Winograd,\\ SQuADv2, ARC-e, ARC-c, CB, ReCoRD} & 85-20\% & -5.8-3.5 \\
        \midrule
        \multirow{2}{*}{\makecell[l]{PALM2\\ \citep{anil2023palm}}}
        & 15 tokens & 0 & \makecell[l]{WikiLingua, XL-Sum, XSum} & 9-47\% & 0.3-0.6 \\
        & 25 tokens & 0 & Proficiency exams & NA & NA \\
        \midrule
        \multirow{3}{*}{\makecell[l]{GPT4\\ \citep{openai2023gpt4}}} 
        & 50 char & 0 & Proficiency exams & <4\% \\
        & 50 char & 0 & \makecell[l]{GSM8k, MMLU, AI2, WinoGrande (1000 samples)} & 0.6-3.4\% & NA \\
        & 50 char & 0 & \makecell[l]{HumanEval, DROP} & 21-25\% & <0\% \\
        \midrule
        \multirow{3}{*}{\makecell[l]{GPT3\\ \citep{brown2020fewshot}}} 
        & 8-13 words & 0 & \makecell[l]{DROP,PIQA } & 89-93\% & 3-7 \\
        & \makecell[c]{8-13 words} & 0 & \makecell[l]{Anagrams 2, SQuADv2, Winograd, RACE-m, CB\\ WMT'16(De→En, En→De),  Anagrams 1} & 3-94\% & 1-2.8 \\
        & \makecell[c]{8-13 words} & 0 & \makecell[l]{ANLI R1, Quac, Symbol Insertion, CoQa, ReCoRD, BoolQ, NQ\\
                                                      MultiRC, RACE-h, LAMBADA, LAMBADA (No Blanks)\\
                                                      WSC, Reversed Words, WMT'16 (En→Ro, Ro→En), \\
                                                      WMT'14 (En→Fr, Fr→En), WebQs, ANLI R2 \\
                                                      TriviaQA, ANLI R3, WiC, RTE, OpenBookQA,\\
                                                      ARC-E, ARC-C, COPA, HellaSwag, Cycled Letters \\
                                                      SAT, Analogies, StoryCloze, Winogrande} & 1-100\% &  -9.8-0.9 \\\\
        \midrule
        \multirow{3}{*}{\makecell[l]{GLaM\\ \citep{Du2021GLaMES}}} 
        & 8-13 words & 0 & \makecell[l]{ ANLI R1, ANLI R2, ANLI R3, ARC-C, ARC-E, \\
        BoolQ, CB, COPA, CoQa, DROP, HellaSwag, \\ 
        LAMBADA, MultiRC, NQ, OpenBookQA, PIQA, \\
        Quac, RACE-h, RACE-m, RTE, ReCoRD, SQuADv2, \\
        StoryCloze, TriviaQA, WSC, WiC, Winograd, Winogrande} & 0-99.66\% &  NA \\\\
        \midrule
        \multirow{3}{*}{\makecell[l]{FLAN\\ \citep{wei2021finetuned}}} 
        & 8-13 words & 0 & \makecell[l]{ ANLI R2} & 97.9\% &  4.8 \\
        & \makecell[c]{8-13 words} & 0 & \makecell[l]{ DROP, ANLI R1, MultiRC, PIQA, ANLI R3, HellaSwag\\
        RTE, WMT’14 (En→Fr, Fr→En), BoolQ, TQA, ARC-E \\
        ARC-C, OpenbookQA, WMT’16(En→De, De→En) \\
        WMT’16(En→Ro, Ro→En), COPA, CB, NQ,  StoryCloze \\
        Winogrande, SQuADv2, ReCoRD} & 0.2-99.4\% &  -10.6-0.5\\ 
        \midrule
        \multirow{3}{*}{\makecell[l]{Gemini Pro \& Ultra\\ \citep{geminiteam2024geminifamilyhighlycapable}}} 
        & NA & 0 & \makecell[l]{ HellaSwag } & NA &  4.9 \& 8.2 \\
        \\
        \\
        \midrule
        \multirow{3}{*}{\makecell[l]{Gemini 1.5\\ \citep{geminiteam2024gemini15unlockingmultimodal}}} 
        & NA & 0 & \makecell[l]{ HumanEval } & 100\% &  14.6 \\
        \\
        \\
        \bottomrule
    \caption{\textbf{Contamination metrics used by different LLM releases, and their (approximate) results}. 
    In this table, we attempt to provide an overview of the various contamination metrics and their parameters used by previous LLM releases, along with the reported results.
    Where possible, we computed EPG from the values reported in the papers. 
    In some cases, the information to do so was not sufficient or particular metrics were not provided. 
    These are marked with NA or presented in the format they were shared in the paper. For the Gemini and Gemini 1.5 model case stuides with HellaSwag and HumanEval are shared in the paper and we present these results with the information available.}\label{tab:lit_overview_table}
\end{longtable}

\normalsize

\section{Benchmarks}
\label{appx:example_prompts}

For the evaluation benchmarks we chose 13 datasets that are widely used for evaluating LLMs from various task types. 
Below, we briefly describe each dataset.
We run the reported benchmarks in two different settings.
For the `generative benchmarks', the model has to generate the correct response given the input prompt.
We run the following benchmarks in generative mode: GSM8K, HumanEval, NaturalQuestions, Deepmind Coding Contest, TriviaQA, BBH, and MBPP. 
A second set of benchmarks are instead ran in `choice' mode.
These benchmarks are ran in a multiple choice setup, where we present the question together with all the possible answers, all represented by a letter (A, B, C, etc.) and ask the model to predict the letter associated with the correct choice.
We extract the model prediction by selecting the letter with the highest probability.
We use this setup for COPA, SiQA, and MMLU.
Lastly, for HellaSwag and PiQA, we extract model answers by comparing the probabilities of the answers rather than the letters, and selecting the highest probability answer.
For these benchmarks, we add an additional analysis that considers whether it matters more whether the contamination is in the question or answer field (for an overview, see \cref{tab:contamination_types}).

\paragraph{COPA}
The Choice of Plausible Alternative (COPA) was presented by \citet{gordon2011copa} as a commonsense reasoning task. 
For each premise there is two choices and a question type that indicates whether the correct choice caused the premise or was result of it(effect). 
There is no specific context or source, and the dataset contains 100 examples in the validation split. 
The model is expected to identify the correct choice based on the question type indicated.  

\paragraph{GMS8K}
GSM8K \citep{gsm8k} is a MATH benchmark consisting of grade school math problems.
The problems are formatted as questions and answers. 
The model is expected to answer the questions correctly through a free-form generation. 
The models are evaluated based on how many open-ended questions they can answer. 
The validation set consists of almost 8.8K questions. 

\paragraph{HellaSwag} 
\citet{zellers2019hellaswag} presented HellaSwag datasets as a commonsense reasoning dataset. 
Given a context can a model choose the correct continuation of a text from 4 choices. 
The incorrect choices are specifically chosen to be challenging options for models that humans can differentiate with the given context. 
The validation and test sets contain little more than 10K examples. 

\paragraph{HumanEval}
HumnEval is a coding benchmark consisting of 164 coding questions released by \citet{humaneval}. 
The dataset gives the model the name of the function and a docstring and expects the model to complete the function correctly. 
The model is expected to generate a passing solution and is evaluated on passing test cases for each question. 

\paragraph{MMLU}
MMLU is a four-way multiple-choice question-answering task presented by \citet{mmlu}. 
The dataset consists of 57 tasks that range from math questions to law and history; questions test knowledge as well as reasoning and consists of almost 16K questions.

\paragraph{SiQA}
Social interaction QA \citep[SiQA,][]{siqa} is designed to evaluate models' reasoning abilities about social situations. 
Each example consists of a context/situation a question and 3 choices. 
The model is expected to choose the correct answer. 
The validation set consists of 1.9K questions. 

\paragraph{PiQA}
PiQA \citep{bisk2019piqa} is a benchmark consisting of questions that test physical reasoning abilities. 
The examples consist of a goal and two choices and the model is expected to reason about the choices and identify which would achieve the given goal. 
The test set consists of close to 1.9K questions. 

\paragraph{Natural Questions}
Natural Questions \citep{naturalquestions} is a dataset of collected questions people asked on the Google search engine. 
The dataset can be formulated as a document question answering task by providing the relevant Wikipedia page but can also be formulated as closed book question answering task. 
In this work, we use the latter formulation so the data only consists of questions the model is expected to generate answers. 
The evaluation is then done with exact match to one of the answers. 
The dataset consists of around 8K examples. 

\paragraph{Deepmind Coding Contest}
Deepmind Coding Contest, or DM Contest \citep{dm_contest} is a code challenge dataset with 165 questions in the test set. 
The inputs define a coding problem with details, unlike HumanEval where the input is a docstring. 
This task presents the model with a natural language description of the problem that needs to be solved and instructions on how to format the output. 
The model is evaluated by whether the generated code passes the test cases. 

\paragraph{TriviaQA}
TriviaQA \citep{joshi-etal-2017-triviaqa} is a large-scale reading comprehension task that requires reasoning over information in a context to answer questions. 
The dataset consists of a context and a question over that context that needs reasoning with multiple pieces of information from that context. 
The outputs are expected to match at least one of the label answers. 

\paragraph{MATH}
The dataset MATH \citet{math} targets mathematical reasoning and consists of competition math problems where the input is a problem description and the generation is parsed and expected to exactly match the correct output. 
The dataset consists of 12.5K examples.   

\paragraph{Big Bench Hard}
Big Bench Hard \citep[BBH][]{bbh} comprises 23 tasks from BIG-Bench \citep{bigbench} where language models had not yet surpassed human annotators at the time of release.
The input is a question and the output is evaluated by whether or not it matches the correct label. 
This subset has around 5K questions that range from solving boolean expressions to sorting words alphabetically. 

\paragraph{MBPP}
Mostly Basic Programming Problems \citep[MBPP]{mbpp} is a 1K dataset of programming questions in Python. 
The inputs are natural language definition of coding problems with a few tests/conditions and the model generated code is executed on test cases. 

\begin{table}[h!]
    \centering
    \resizebox{\textwidth}{!}{
    \renewcommand{\arraystretch}{1.4} 
    \begin{tabular}{lll}
        \toprule
        \textbf{Type} & \textbf{Definition} & \textbf{Examples} \\
        \midrule
        Question & \makecell[l]{Contamination that is bound in the question/prompt field\\ 
        and does not contain any tokens from the correct choice \\ when the contamination is extended to the fullest} & \makecell[l]{holding a pocket knife while } \\
        \midrule
        Answer & \makecell[l]{Contamination that is bound in the correct choice field and\\
    does not contain any tokens from the question when the \\contamination is extended to the fullest} & \makecell[l]{a small stone from the flowing river} \\
        \midrule
        Full prompt & \makecell[l]{Contamination can be anywhere in the full prompt, \\
        including in just the question or answer. We take the \\ 
        longest contaminating subsequence}. & \makecell[l]{in the wilderness. Then he takes a small stone\vspace{1mm} \\ a small stone from the flowing river \vspace{1mm} \\ holding a pocket knife while } \\
        \midrule
        \makecell[l]{Across question \& \\answer boundary} & \makecell[l]{Contamination spans over the question and answer boundary \\
        in the fully prompted example. The \texttt{skipgram} method can \\
        use a \texttt{skip\_budget} to skip a limited number of tokens, but \\
        contamination must match at least 1 token in both fields} & \makecell[l]{in the wilderness. Then he takes a small stone} \\
    \bottomrule
    \end{tabular}
    }
    \caption{\textbf{Different types of contamination `spans'.}
    Given the Question `\textit{A man is holding a pocket knife while sitting on some rocks in the wilderness. then he}' and the answer `\textit{takes a small stone from the flowing river and smashes it on another stone}', we give examples of the different types of contamination we find with the \skipgram method. }
    \label{tab:contamination_types}
\end{table}

\section{Contamination metrics -- brief implementation details}\label{appendix:implementation_details}

In this section, we provide brief implementation details for the methods we used.
For further details, please reach out to the corresponding authors.

\paragraph{\openai} To quantify contamination, the \openai method considers the presence of $n$-grams from an evaluation sample in pre-training data. 
In our implementation, we tokenise each training corpus or benchmark and compute hashes for every $n$-gram in it using a polynomial rolling hash function \citep{karp1987efficientrp}.
We then intersect the hash tables of a training corpus and a benchmark to identify matches. We repeat this procedure for each choice of $n\in\{8, 10, 13, 20\}$. To control for $mincount$, we only consider hash values on the training corpus side with the given minimum number of occurrences.

\paragraph{\ngram} The \ngram method uses the same normalisation and rolling hash function from as the \openai method. 
The main difference between the two methods is how we calculate the scores. 
While the \openai method focuses on the number of matching $n$-grams, the \ngram method looks at the total number of tokens that are present in the matching windows. 
The difference can be seen if we consider a sample with two disjoint 8-grams that match. 
For the \openai method, since only two 8-grams match, the nominator of the contamination score would be 2; for the \ngram method there are 16 tokens within the matching spans. 
Thus, the \openai method gives higher scores for longer contiguous matches, while the \ngram method gives higher scores compared to \openai when there are disjoint matches over a text.

\paragraph{\union}
First used by \citet{touvron2023llama2}, the \skipgram method focuses on the longest token span from an evaluation sample which appears in pre-training data, allowing for minor syntactic and formatting differences with a small $skip\_budget$. 
We implement this skipgram-matching procedure by first identifying small seed matches (of size $n=8$ tokens), and then extending these matches to the left and right
(similar to the BLAST algorithm in genetics \citep{blast}) 
while allowing for up to $skip\_budget$ token mismatches. 
In our experiments, we range over different $skip\_budget$ values.
We compute results separately for values of $n>8$, but compute these results post-hoc.

\paragraph{\skipgram} We computed the contamination scores for the \skipgram method are computed ad-hoc from the hashed matches computed for the \union metric, but considering only the longest matching substring, rather than all substrings.


\section{Scaling behaviours}

\begin{figure}[h]
\centering
\input{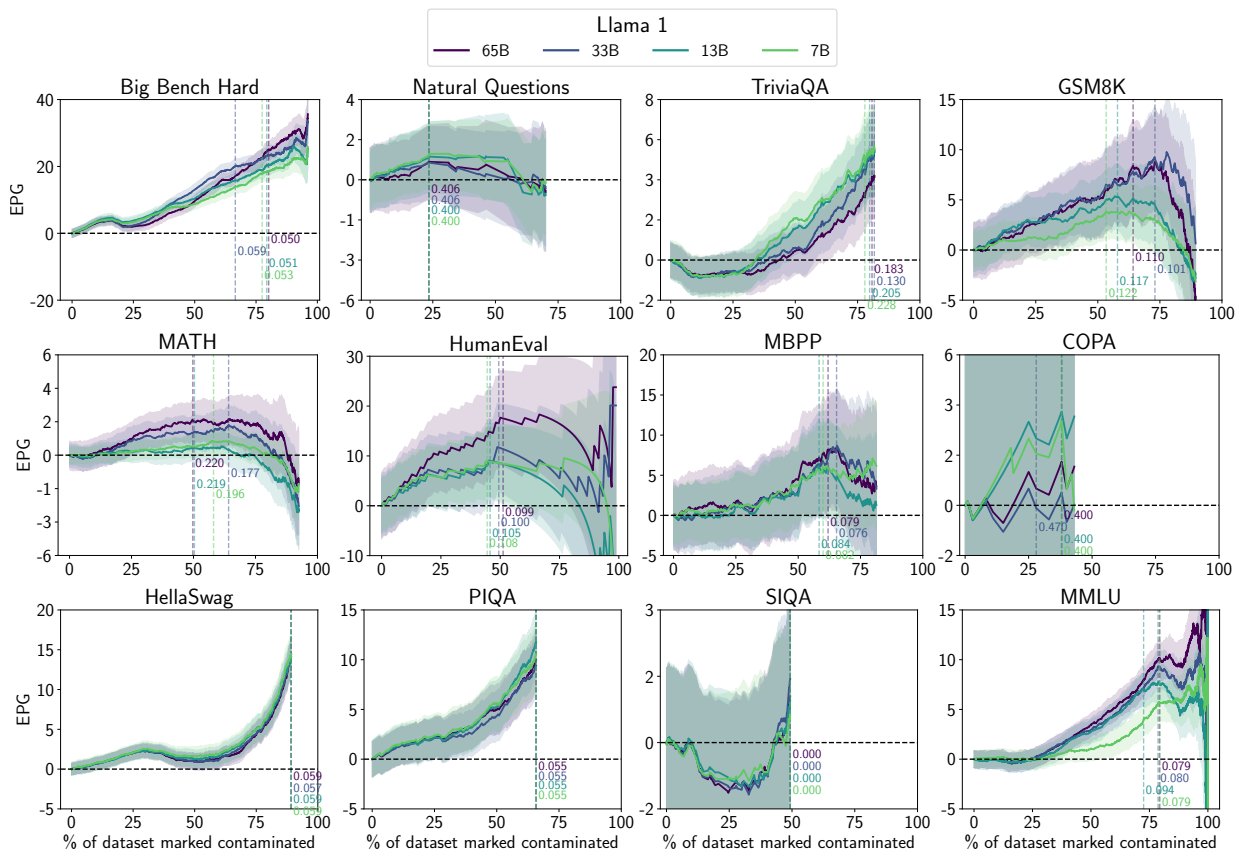}
\caption{\textbf{Scaling behaviour for Llama 1 models.} We show how the effects of contamination change with model size, for Llama 1 models. These plots supplement the analysis presented in \cref{subsec:scaling}. Plots for Pythia models can be found in \cref{fig:scaling_pythia_all}.}
\label{fig:scaling_llama_all}
\end{figure}

\begin{figure}[h]
\centering
\input{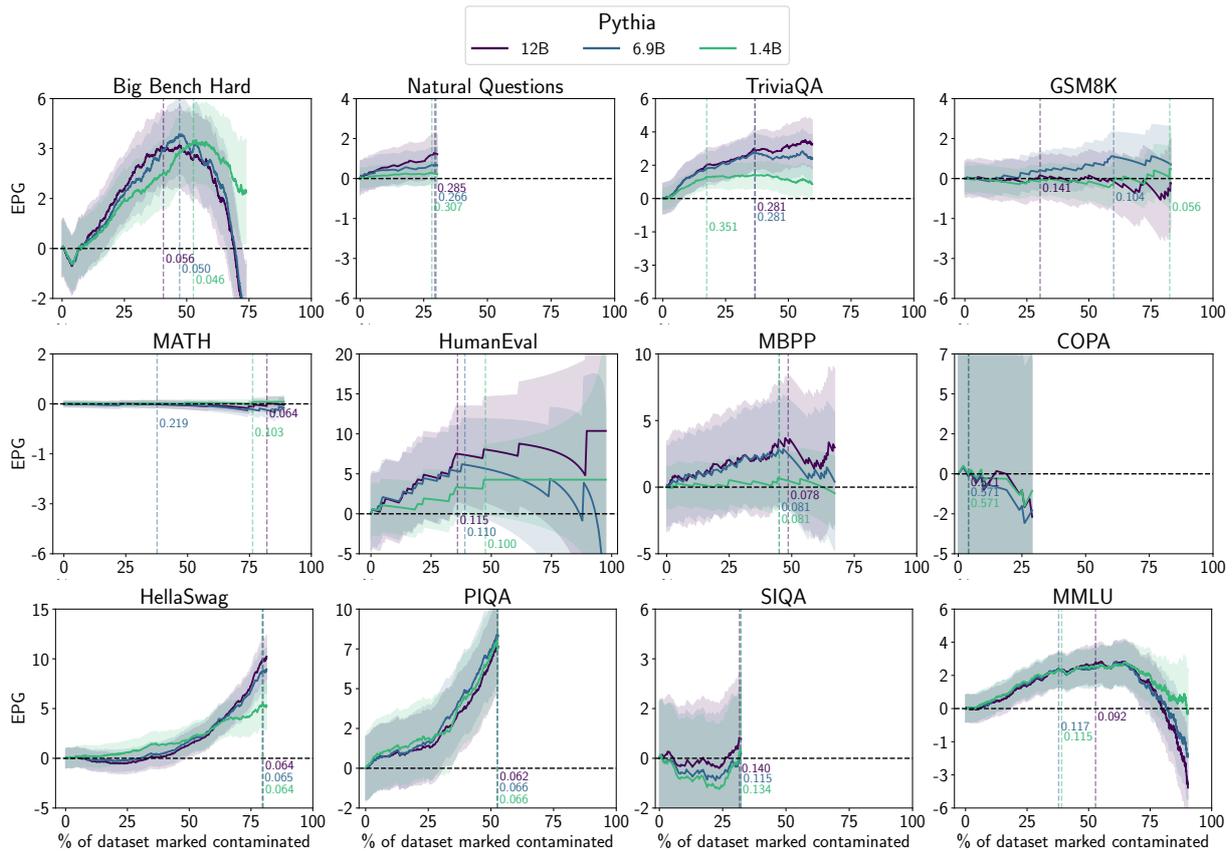}
\caption{\textbf{Scaling behaviour for Pythia models.} We show how the effects of contamination change with model size, for Pythia models. These plots supplement the analysis presented in \cref{subsec:scaling}. Plots for Llama 1 models can be found in \cref{fig:scaling_llama_all}.}
\label{fig:scaling_pythia_all}
\end{figure}

In \cref{subsec:scaling}, we discussed the effect of scale on the impact of contamination, with a few representative examples.
For completeness, we provide the same plots for all benchmarks and both model series here, in \cref{fig:scaling_llama_all} and \cref{fig:scaling_pythia_all}.

\section{Additional plots}

\begin{figure}
    \centering
    \begin{subfigure}[b]{0.59\textwidth}
        \includegraphics[height=5.9cm]{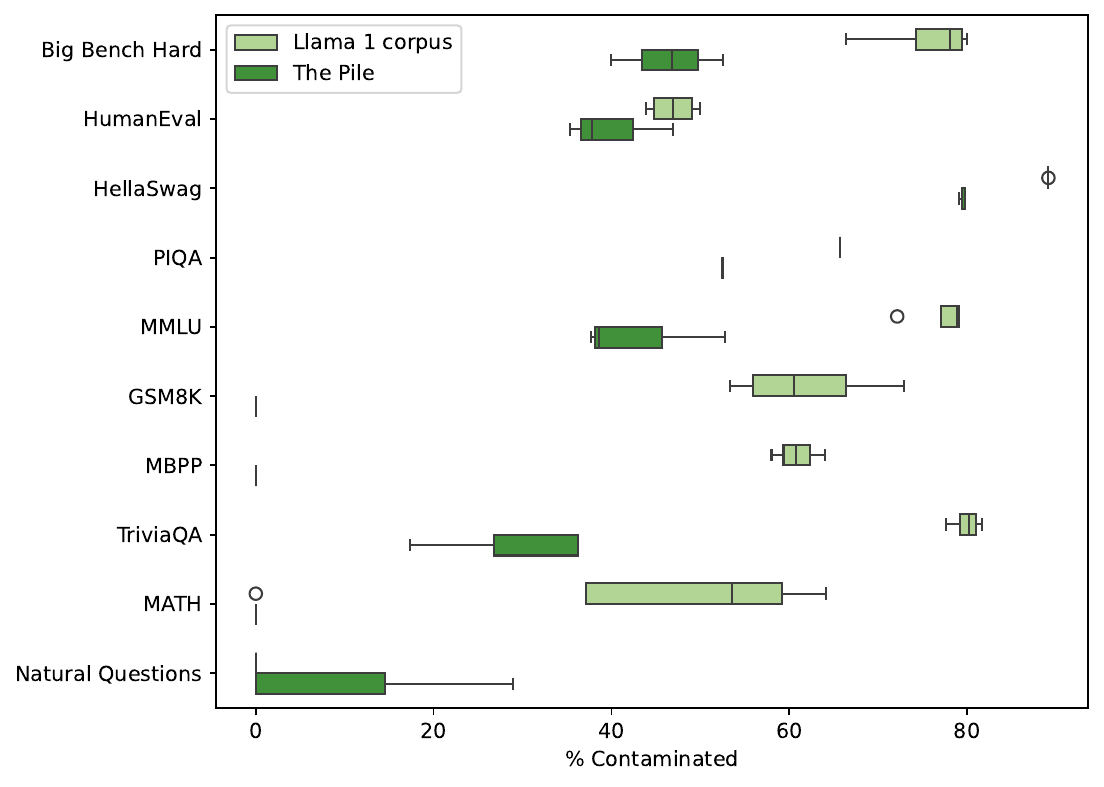}
        \caption{\% Contaminated}
        \label{fig:skip_pct_contam_dist}
    \end{subfigure}
    \hspace{-30pt}
    \begin{subfigure}[b]{0.45\textwidth}
        \includegraphics[height=5.9cm, trim=35mm 0mm 0mm 0mm, clip]{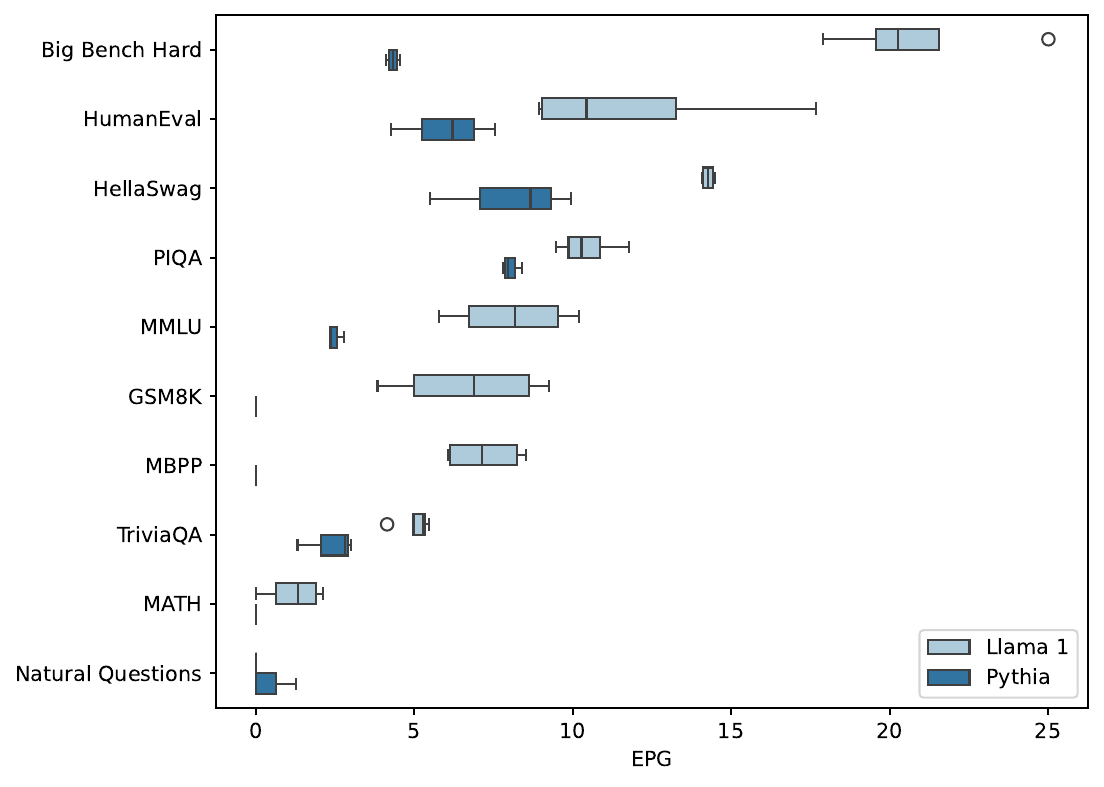}
    \caption{EPG}
    \label{fig:skip_epg_dist}
    \end{subfigure}
    \caption{\textbf{\%Contaminated and EPG across models and benchmarks.} (a) Percentage of the dataset marked contaminated. (b) Corresponding EPG of the model-benchmark pairs we considered in our study, according to the \skipgram method. Optimal thresholds are selected separately for each model-benchmark pair. Per-model values are reported in \cref{fig:skip_epg_pct_contam_all}.}
\label{fig:skip_epg_pct_contam_dist}
\end{figure}

\begin{figure}
    \centering
    \includegraphics[width=1\textwidth]{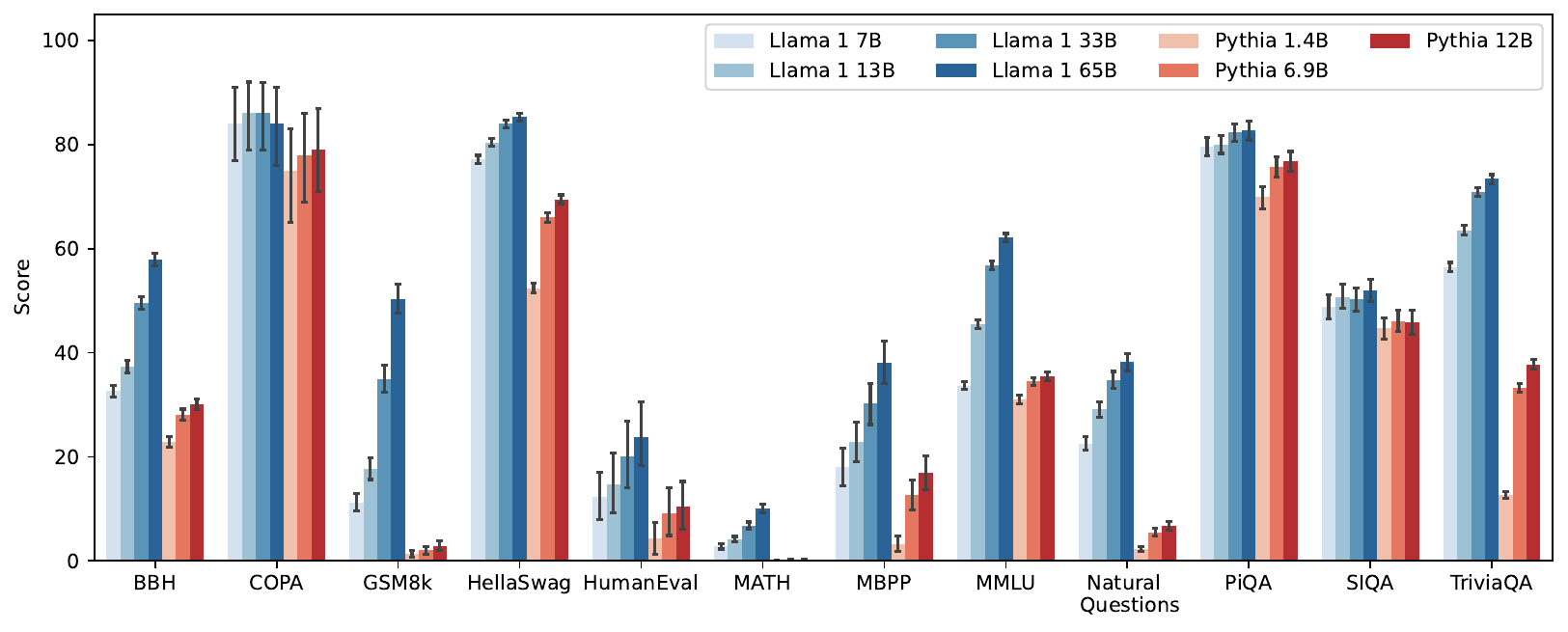}
    \caption{\textbf{Performances across models and datasets.} For context, we show the performances of all models on the benchmarks we consider. Error bars correspond to 95\% CIs.
    }
    \label{fig:all_performances}
\end{figure}

In \cref{fig:skip_epg_pct_contam_dist}, we include box-plots of the percentage marked contaminated and corresponding EPG.
This plot supplements our analysis in \cref{subsec:how_contaminated}.
In \cref{fig:all_performances}, we report the benchmark scores for all models, contextualising our results in \cref{subsec:scaling}.

\section{Extensive contamination analyses}
\label{appx:additional_exp}

\begin{figure*}[h]
    \centering
    \includegraphics[width=\textwidth]{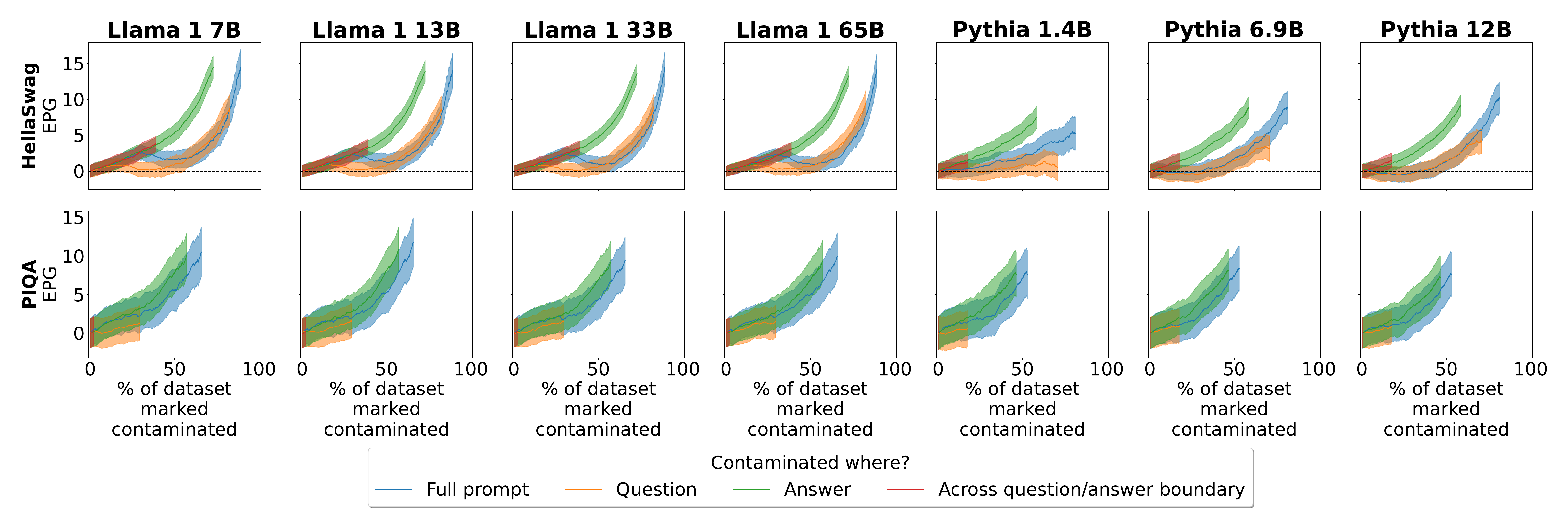}
    \vspace{-1.8em}
    \caption{\textbf{Contam. analysis for different locations of contamination.} For Hellaswag and PIQA we restrict the \plotmethodname to \textit{where} the contamination occurs when using the \skipgram method (with $skip\_budget=0$). Specifically, we see for these benchmarks that restricting to contamination only in the answer leads to qualitatively higher curves, indicating a better ordering of data points. This supports the hypothesis in the main text (\cref{subsec:qualitative_results}, \cref{tab:optimal_thresh}), where we suggested that \skipgram does not perform as well on these benchmarks since it may not ``count'' contamination in only the answer as strongly as \ngram.}
    \label{fig:appx_comp_n_where}
\end{figure*}

\begin{figure*}[ht]
    \centering
    \includegraphics[width=\textwidth]{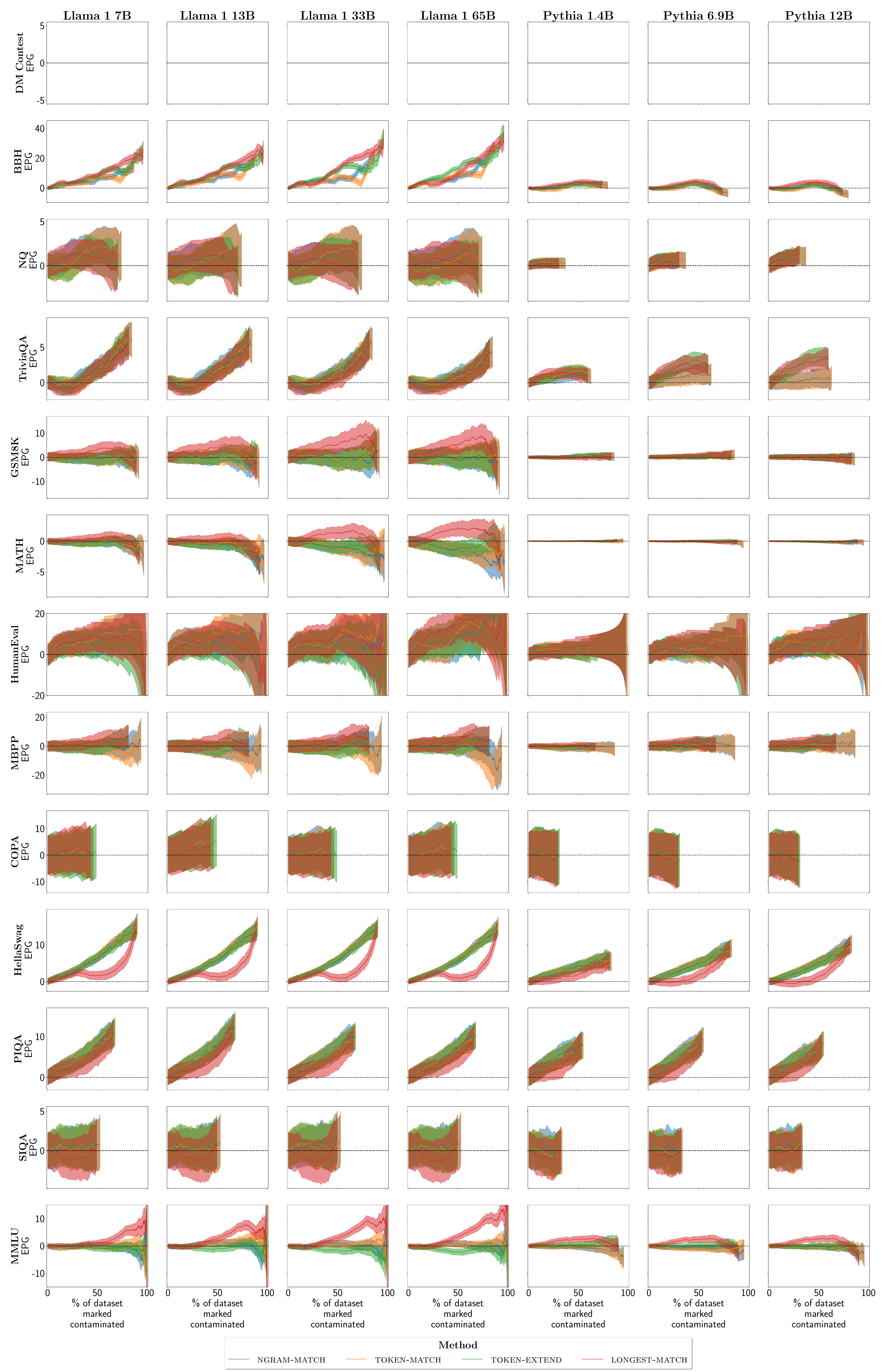}
    \caption{\textbf{Contam. analysis over all benchmarks (rows) and models (columns) as we vary the scoring method.}}
    \label{fig:appx_comp_method}
\end{figure*}

In \cref{sec:analysis}, we discussed how different methods (and hyperparameters settings) may lead to better or worse quantification of contamination. 
Specifically, we focus on interpreting methods in terms of how many true postives versus false positives they add.
To ensure our claims are justified, we conducted analyses across 7 models and 13 benchmarks. 
Due to limited space in the main text, we chose not to include 91 plots per comparison and instead focused on the overarching themes and takeaways. 
Here, we provide these extended results as a resource for future contamination researchers.
First, in \cref{fig:appx_comp_method}, we show how the different methods compare for all benchmarks we considered, given their optimal hyperparameters ($n$=8, $skip$=0 and $mincount$=1).
In the consecutive plots, we include \plotmethodname plots to illustrate how the value of $n$ impacts the found contamination and EPG, for both \openai and \skipgram, respectively; what is the effect of $mincount$ for \openai; and what is the effect of the $skip\_budget$ for \union.
Lastly, we show, for HellaSwag and PiQA, how it matters \emph{where} in the example (the question or the response) the contamination occurs, in \cref{fig:appx_comp_n_where}.
\cref{tab:appx_comp_skip} furthermore shows the average and maximum changes given different skip budgets, supplementing our analysis in \cref{subsec:skip_budget}.

\begin{figure*}[ht]
    \centering
    \includegraphics[width=\textwidth]{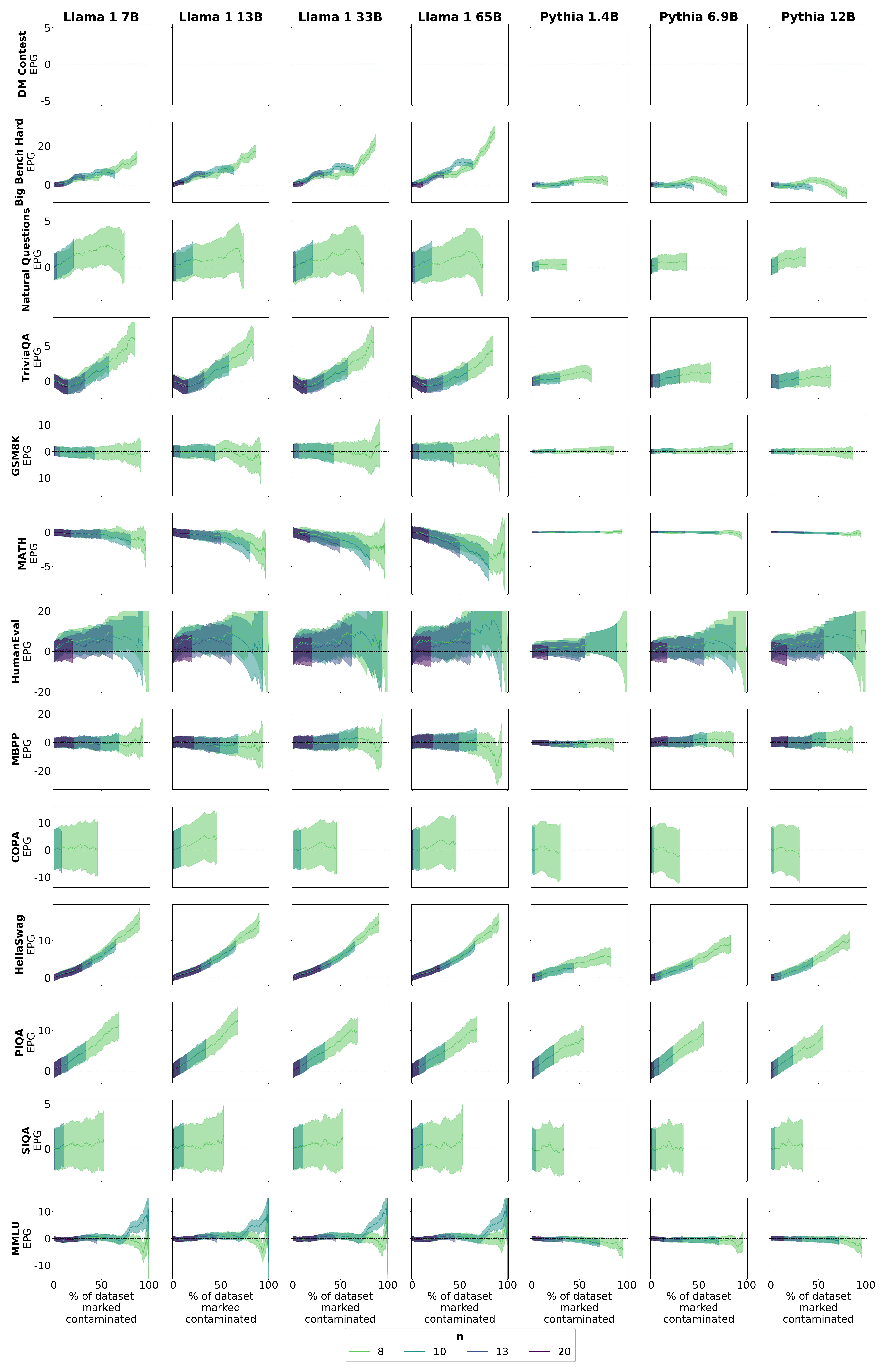}
    \vspace{-2.5em}
    \caption{\textbf{Contam. analysis over all benchmarks (rows) and models (columns) as we vary $n$ in \openai.}}
    \label{fig:appx_comp_n}
\end{figure*}

\begin{figure*}[ht]
    \centering
    \includegraphics[width=\textwidth]{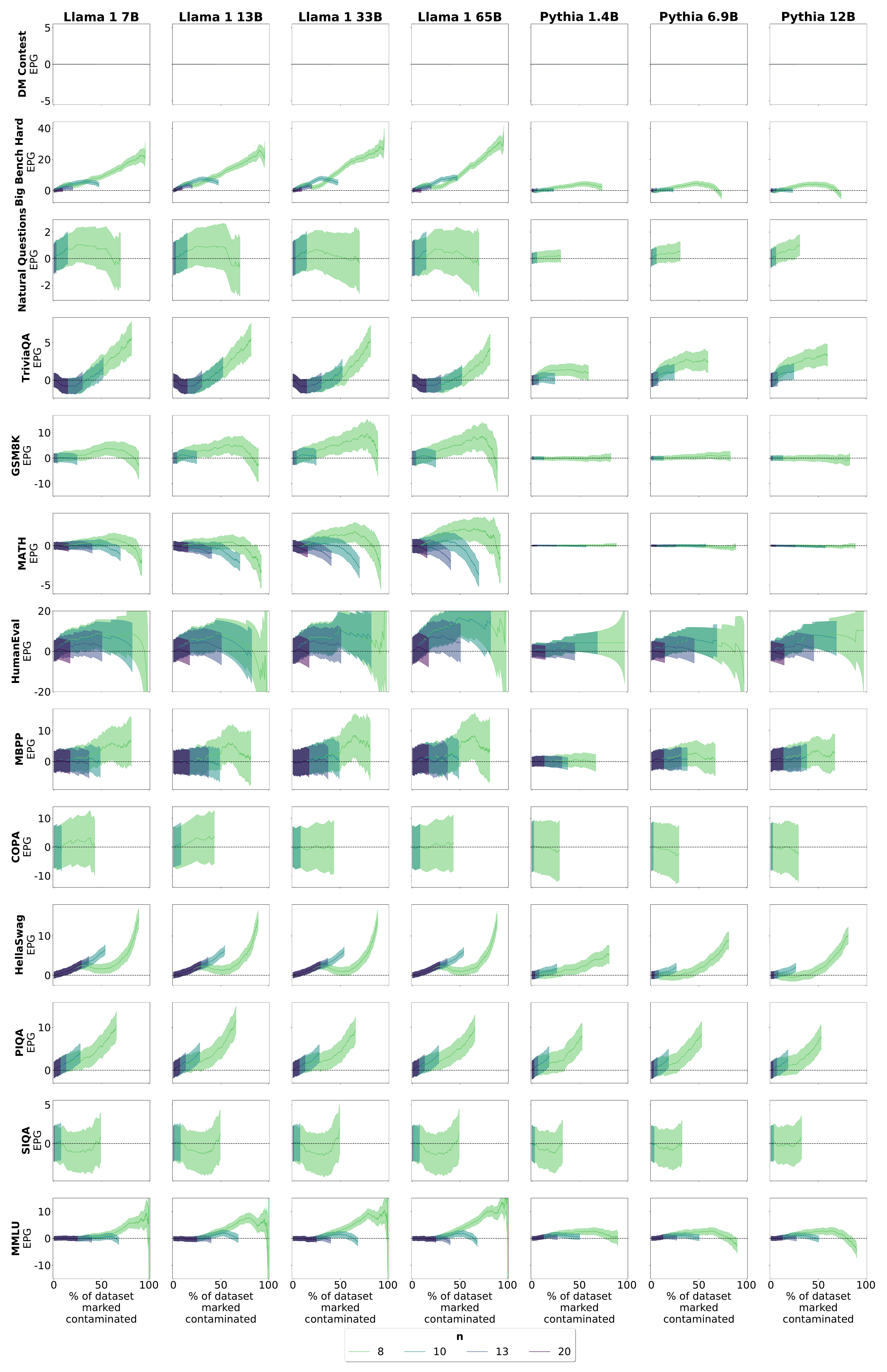}
    \vspace{-2.5em}
    \caption{\textbf{Contam. analysis over all benchmarks (rows) and models (columns) as we vary $n$ in \skipgram.}}
    \label{fig:appx_comp_n_skipgram}
\end{figure*}

\begin{figure*}[ht]
    \centering
    \includegraphics[width=\textwidth]{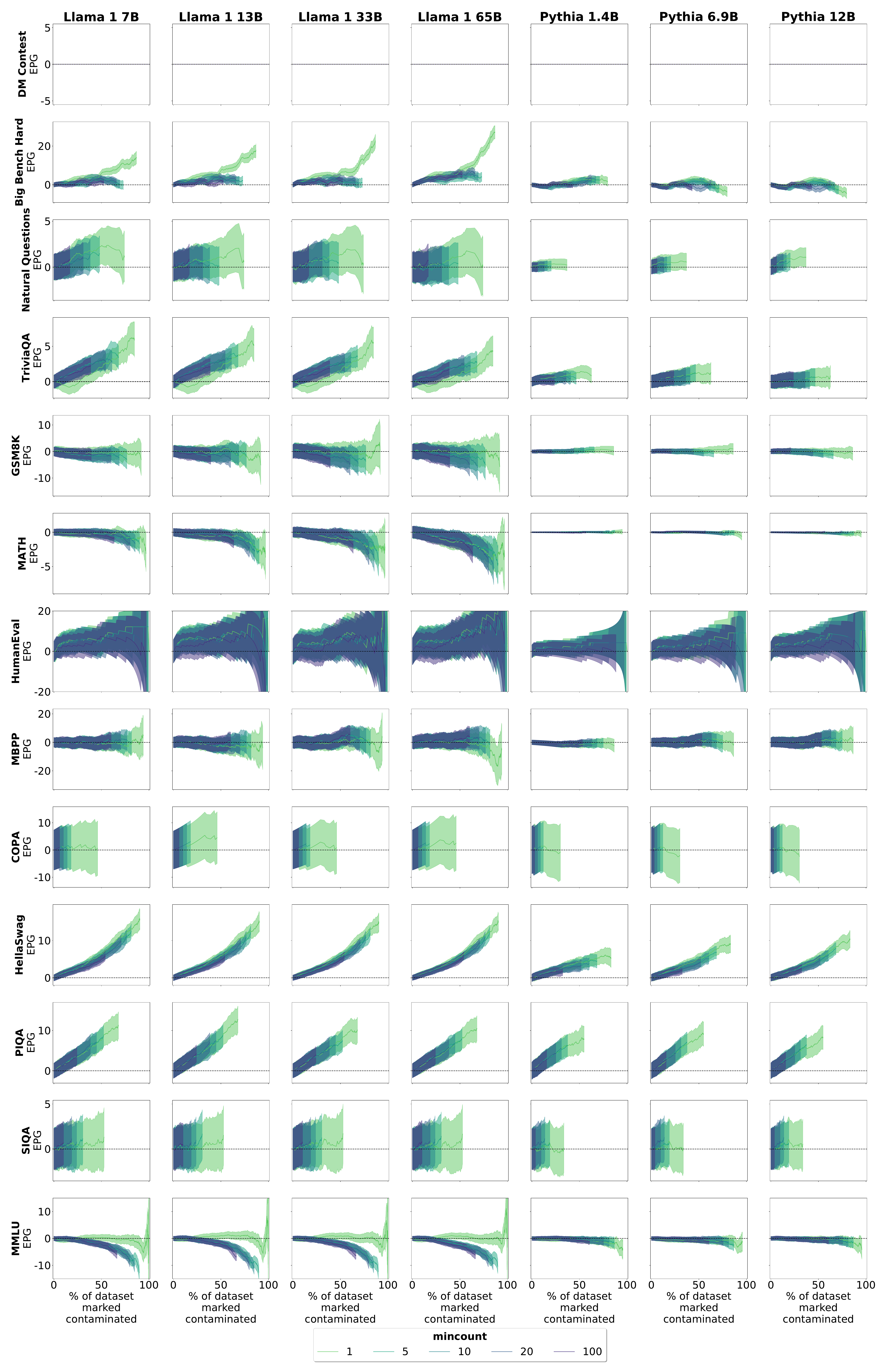}
    \vspace{-2.5em}
    \caption{\textbf{Contam. analysis over all benchmarks (rows) and models (columns) as we vary $mincount$ in \openai.}}
    \label{fig:appx_comp_mincount}
\end{figure*}

\begin{figure*}[ht]
    \centering
    \includegraphics[width=\textwidth]{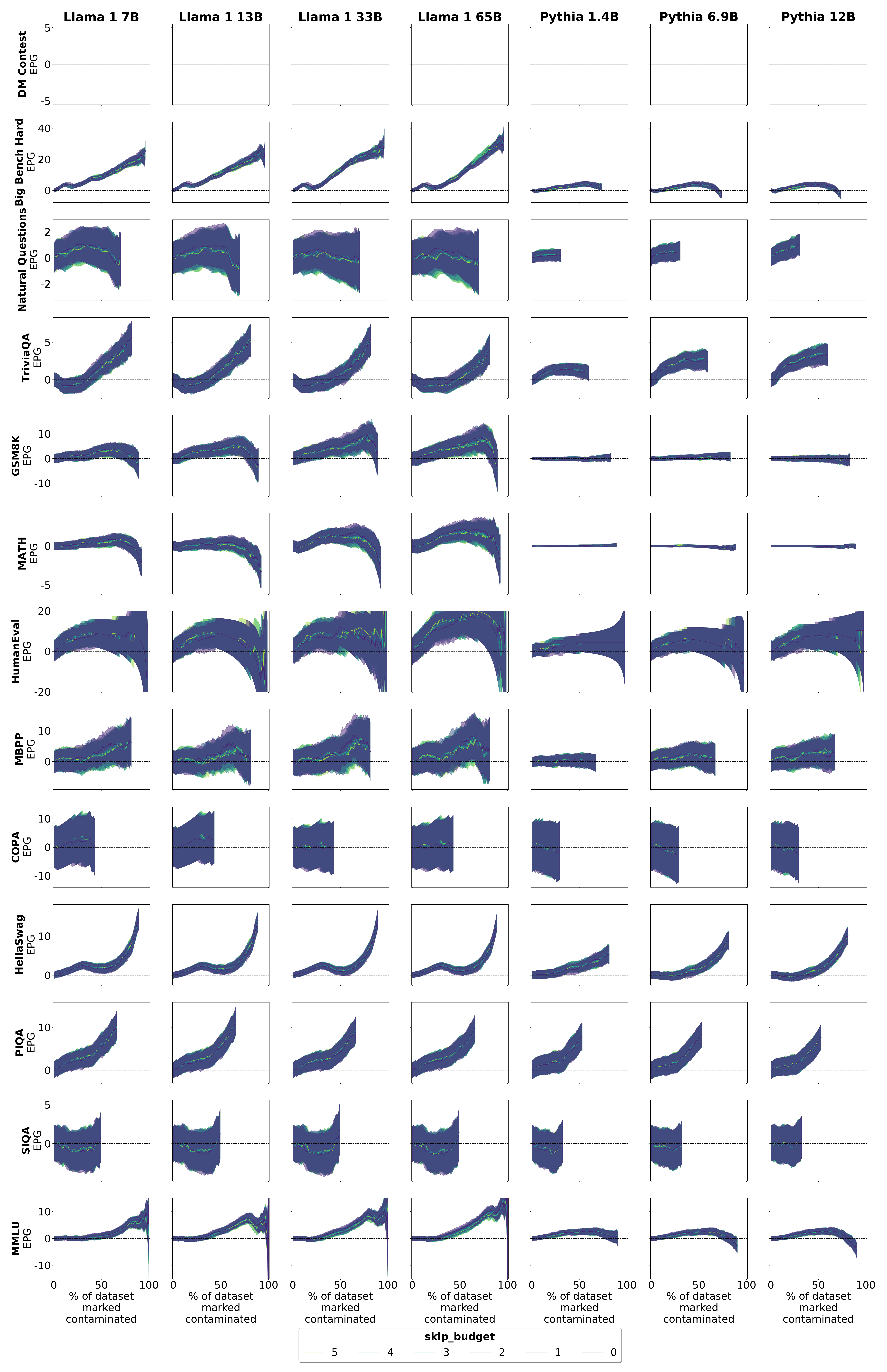}
    \vspace{-2.5em}
    \caption{\textbf{Contam. analysis over all benchmarks (rows) and models (columns) as we vary $skip\_budget$ in \union.}}
    \label{fig:appx_comp_skip}
\end{figure*}

\begin{table}[H]
    \centering
    \resizebox{0.45\columnwidth}{!}{
    \begin{tabular}{lcccc}
        \toprule
        \multirow{2}{*}{Benchmark} & \multicolumn{2}{c}{Llama 1 Corpus} & \multicolumn{2}{c}{The Pile} \\
        & mean & max & mean & max \\ \midrule
        DM Contest & 0.059 & 0.180 & 0.053 & 0.155 \\
        Big Bench Hard & 0.033 & 0.225 & 0.023 & 0.271 \\
        Natural Questions & 0.023 & 0.333 & 0.008 & 0.300 \\
        TriviaQA & 0.032 & 0.526 & 0.023 & 0.360 \\
        GSM8K & 0.031 & 0.386 & 0.017 & 0.289 \\
        MATH & 0.048 & 0.540 & 0.044 & 0.351 \\
        HumanEval & 0.059 & 0.255 & 0.052 & 0.197 \\
        MBPP & 0.028 & 0.214 & 0.025 & 0.214 \\
        COPA & 0.012 & 0.211 & 0.009 & 0.176 \\
        HellaSwag & 0.028 & 0.424 & 0.017 & 0.448 \\
        PIQA & 0.018 & 0.333 & 0.013 & 0.312 \\
        SIQA & 0.009 & 0.235 & 0.004 & 0.156 \\
        MMLU & 0.031 & 0.463 & 0.025 & 0.258 \\ \bottomrule
    \end{tabular}
    }
    \caption{\textbf{Average and maximum changes between $skip\_budget$ 0 and 5.} We show the average and maximum changes in contamination scores using the \union method when allowing a $skip\_budget$ of 5 tokens (versus 0). The minimum change is 0, since $skip\_budget$ of 5 is a less strict method (finding at least as many contaminated tokens per evaluation sample).}
    \label{tab:appx_comp_skip}
\end{table}


\end{document}